\documentclass[journal,twoside,web]{ieeecolor}
\usepackage{generic}
\usepackage{cite}
\usepackage{amsmath,amssymb,amsfonts}
\usepackage{algorithm}
\usepackage{graphicx}
\usepackage{textcomp}
\usepackage{subfigure}
\usepackage{algpseudocode}
\usepackage{epsfig}
\usepackage{color}
\usepackage{multirow}
\usepackage{makecell}
\usepackage{booktabs}
\usepackage{mathrsfs}
\usepackage{indentfirst}
\usepackage{url}
\usepackage{xspace}

\newcommand{\method}{\textit{SPSSOT}\xspace}

\def\BibTeX{{\rm B\kern-.05em{\sc i\kern-.025em b}\kern-.08em
    T\kern-.1667em\lower.7ex\hbox{E}\kern-.125emX}}
\begin{document}

\title{Cross-hospital Sepsis Early Detection via Semi-supervised Optimal Transport with Self-paced Ensemble}
\author{Ruiqing~Ding,
        Yu~Zhou,
        Jie~Xu,
        Yan~Xie,
        Qiqiang~Liang,
        He~Ren,\\
        Yixuan~Wang,
        Yanlin~Chen,
        Leye~Wang
        and~Man~Huang
\thanks{Manuscript received 6 December 2021; revised 8 October 2022; accepted 28 February 2023. 
This work was supported by grants from the National Natural Science Foundation of China (No. 81801940 Y.Z., 61972008 L.W.). }
\thanks{Ruiqing Ding and Leye Wang are with Key Lab of High Confidence Software Technologies (Peking University), Ministry of Education, China, and also with School of Computer Science, Peking University, Beijing 100871, China (e-mail: ruiqing@stu.pku.edu.cn, leyewang@pku.edu.cn).}
\thanks{Yu Zhou, Qiqiang Liang and Man Huang are with General Intensive Care Unit, Zhejiang University School of Medicine Second Affiliated Hospital, Hangzhou 310009, Zhejiang, China (e-mail:\{naseph, deter\_leung, huangman\}@zju.edu.can).}
\thanks{Jie Xu is with IT center, Zhejiang University School of Medicine Second Affiliated Hospital, Hangzhou 310009, Zhejiang, China (e-mail:2202113@zju.edu.cn).}
\thanks{Yan Xie and He Ren are with Beijing HealSci Technology Co., Ltd., Beijing 100176, China (e-mail: \{yan.xie, he.ren\}@healscitech.com).}
\thanks{Yixuan Wang is with Department of Computer Science and Technology, Peking University, Beijing 100871, China (e-mail: kugamashiro@pku.edu.cn).}
\thanks{Yanlin Chen is with School of Information Management and Engineering, Shanghai University of Finance and Economics, Shanghai 200433, China (e-mail: 1391469597@qq.com).}
\thanks{(Equal Contribution: Ruiqing Ding and Yu Zhou, Corresponding author: Man Huang.)}
}

\maketitle
\begin{abstract}
Leveraging machine learning techniques for Sepsis early detection and diagnosis has attracted increasing interest in recent years. 
However, most existing methods require a large amount of labeled training data, which may not be available for a target hospital that deploys a new Sepsis detection system. More seriously, as treated patients are diversified between hospitals, directly applying a model trained on other hospitals may not achieve good performance for the target hospital. To address this issue,  
we propose a novel semi-supervised transfer learning framework based on optimal transport theory and self-paced ensemble for Sepsis early detection, called \method, which can efficiently transfer knowledge from the source hospital (with rich labeled data) to the target hospital (with scarce labeled data). 
Specifically, \method incorporates a new optimal transport-based semi-supervised domain adaptation component that can effectively exploit all the unlabeled data in the target hospital. 
Moreover, self-paced ensemble is adapted in \method to alleviate the class imbalance issue during transfer learning.
In a nutshell, \method is an end-to-end transfer learning method that automatically selects suitable samples from two domains (hospitals) respectively and aligns their feature spaces. 
Extensive experiments on two open clinical datasets, MIMIC-III and Challenge, demonstrate that \method  outperforms state-of-the-art transfer learning methods by improving 1-3\% of AUC.
\end{abstract}


\section{Introduction}
\label{sec:introduction}
\IEEEPARstart{S}{epsis}  is a life-threatening disease that occurs when the body's response to infection is out of balance \cite{10.1001/jama.2016.0287}. 
In severe cases, it will trigger body changes that may damage multiple organ systems and lead to death \cite{zimmerman2015pediatric}. 
Sepsis has become a major cause of in-hospital death for intensive care unit (ICU) patients, which places an enormous burden on public health expenditures \cite{fleischmann2016assessment} \cite{ou2017impact}.  
In 2013, Sepsis was responsible for 10\% of the ICU admissions and occupied about 25\% of the ICU beds in US hospitals, accounting for over \$23.6 billion (6.2\%) of total US hospital costs  \cite{sheetrit2017temporal}. 
Early detection is crucial to the sepsis management; with each one-hour delay in the administration of antibiotic treatment, the mortality rate increases by 7\% \cite{kumar2006duration}. 

Recently, machine learning techniques start to be applied in Sepsis diagnosis and early detection, such as the linear model\cite{shashikumar2017early}, Support Vector Machine \cite{horng2017creating}, Neural Network \cite{futoma2017learning}, Gradient Boosting Decision Tree \cite{li2020time}.
These methods need huge amounts of labeled training data to ensure performance. 
In reality, one hospital may hold its treated patients' Electronic Medical Records (EMRs), but it is common that most EMRs are not properly labeled for a machine learning task (e.g., Sepsis early detection) \cite{fleuren2020machine}. 
Therefore, how to use these (unlabeled) records to predict the health situations of new patients is an important problem to be addressed. 

\textit{Transfer learning} \cite{pan2009survey} is a promising machine learning paradigm for the label-scarcity scenario; 
it provides an unconventional perspective to transfer external knowledge from another hospital with rich labeled data to improve the machine learning performance of a target hospital with scarce labeled data. 
It also reduces expensive data-labeling costs. 
The state-of-the-art transfer learning strategy is fine-tuning \cite{lee2012adapting} \cite{choi2016doctor}; however, overfitting is often caused by fine-tuning a large number of parameters with very small labeled data \cite{gupta2020transfer}. 
Consequently, many challenges still exist for successful knowledge transfer, especially in clinical data:

\textbf{Covariate Shift.} Different medical devices in different hospitals may result in diverse test values.
Also, patients' agglomeration factor cannot be neglected. 
Specifically,  patients tend to choose a hospital that is more appropriate for their diseases and health conditions \cite{luft1990does}. 
Hence, the sets of patients' information collected from two hospitals are often different from each other; in other words, they are not in the same feature space. 
Consequently, it is essential to map them into a common hidden space, known as \textit{domain adaptation} \cite{pan2010domain}.

\textbf{Label shift.} Label shift implies that the label distribution changes from the source to the target\cite{azizzadenesheli2019regularized}. 
In particular, the incidence of a disease may fluctuate with locations and time, easily leading to a negative transfer. 
To alleviate this pitfall, prior methods propose to re-weight source samples' importance \cite{cortes2010learning}; however, they incur a huge computational burden when a large number of samples exist.

\textbf{Class Imbalance.} Imbalanced data is ubiquitous especially for medical diagnostic datasets \cite{yang2020rethinking}, and it exhibits a long-tailed distribution\cite{buda2018systematic}. 
During transfer learning, it is also vital to reduce the classification bias caused by data imbalance and find more appropriate decision boundaries.

\begin{figure} 
    \centering
    \includegraphics[width=1.0\linewidth]{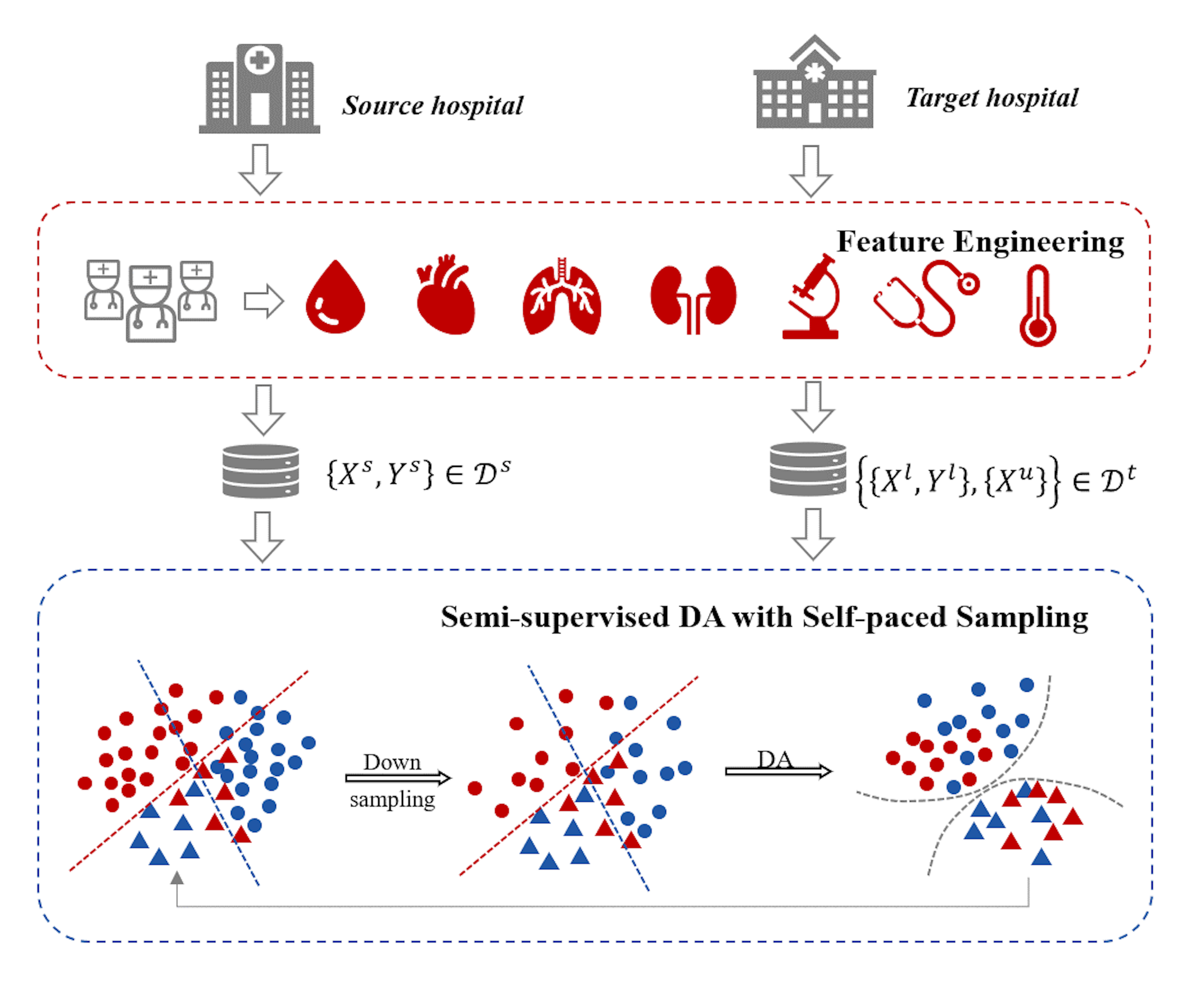}
    \caption{The Overall Framework of Semi-supervised Optimal Transport with Self-paced Ensemble (\method), which consists of 3 main parts: (1) Feature Engineering to extract Sepsis-related features under the guidance of doctors; (2) Self-paced sampling to filter out ``more contributing" samples from negative samples (no Sepsis, shown as circles), which will help to improve the performance of classifier. 
    (3) DA (Semi-supervised domain adaptation): given all data with labels of source hospital (shown as red), little data with labels and most data without labels in target hospital (shown as blue), align two feature spaces via optimal transport theory and learn a better classifier to distinguish whether Sepsis will occur. \textit{Best viewed in color}.}
    \label{fig:overall}
\end{figure}

To overcome the above difficulties, we propose a semi-supervised transfer learning approach based on optimal transport \cite{villani2008optimal} and self-paced ensemble \cite{liu2020self} approach to complete cross-hospital Sepsis early detection. 
There are \textbf{three main components} of \method: \textit{feature engineering} under the guidance of doctors to extract features associated with Sepsis, \textit{self-paced ensemble} to achieve data balance, and \textit{semi-supervised domain adaptation via optimal transport} to tackle the problem of inconsistent feature spaces. 
The overall framework is shown in Fig. \ref{fig:overall}. Our contributions are summarized as follows:
 
(1) To the best of our knowledge, this is the first work on cross-hospital Sepsis early detection. In particular, by properly transferring knowledge from another hospital with rich labeled data, our method can enable good detection performance  for the target hospital with little labeled data.

(2) Considering the inconsistent feature distributions and the unbalanced noisy data status in cross-hospital Sepsis early detection, we propose a novel end-to-end deep transfer learning framework, called \method, consisting of three components: feature engineering, semi-supervised domain adaptation with optimal transport, and self-paced ensemble. 
More specifically, in \textit{semi-supervised domain adaptation with optimal transport}, we design a label-adaptive optimal transport strategy to achieve the precise-pair-wise optimal transport, and an intra-domain deep feature  discrimination strategy to find a better decision boundary. 
In \textit{self-paced ensemble}, we improve and incorporate an ensemble algorithm for imbalanced classification \cite{liu2020self} into our deep transfer learning framework, which can adaptively downsample the majority data from both domains to alleviate the class imbalance issue.

(3) By conducting the experiments on mutual transfer between two open clinical datasets, MIMIC and Challenge, we have validated \method to improve the AUC values by at least 3\% and 1\% with only 1\% labeled data in the target domain compared to state-of-the-art transfer learning methods \cite{saito2019semi,9578015,Yoon_2022_WACV}.

\section{Related Work}
This paper mainly provide a new solution for Sepsis early detection when there are few labeled EMRs in the hospital. 
We propose a transfer learning framework based on optimal transport theory  \cite{monge1781memoire} to cope with the data discrepancy between different hospitals; and introduce a self-paced ensemble method to overcome the extreme label imbalance problem.
Accordingly, we provide a brief overview of related work in four fields, i.e., Sepsis early detection, transfer learning with optimal transport, data imbalance, and self-paced learning.

\textbf{Sepsis Early Detection.}
Machine learning (ML) techniques excel in the analysis of complex signals in data-rich environments which promise to improve the early detection of Sepsis.
Most of the studies are carried out in the ICU \cite{fleuren2020machine} \cite{karpatne2018machine}, and some of them are specifically on neonatal Sepsis \cite{8758174} \cite{9186288}. 
Systematic review and meta-analysis indicate that individual machine learning models can accurately predict the onset of Sepsis in advance on retrospective data \cite{li2020time} \cite{8624374}. 
The PhysioNet/Computing in Cardiology (CinC) Challenge 2019 focused on this issue and promoted the development of open-source AI algorithms for real-time and early detection of Sepsis \cite{reyna2019early}.
Such approaches, which typically apply ML techniques to clinical data, can dynamically suggest real-time predictions and optimal treatments for Sepsis patients and yield excellent results in the medical field. 
However, the variety of studies engaged in Sepsis early detection without sufficient labeled data remains small.

\textbf{Transfer Learning with Optimal Transport.}
The core of transfer learning is to align the source and target distributions by minimizing a divergence that measures the discrepancy between them. 
Optimal Transport (OT) theory can be regarded as one of the discrepancy-based alignment methods, as it can be used for calculating Wasserstein distances between probability distributions \cite{courty2016optimal}. 
Given the cost function (e.g., $l_2$ distance) between samples in the source and target domains, we can calculate the probabilistic coupling matrix $\gamma$.  
It has been applied in domain adaptation to learn the transformation between different domains \cite{courty2014domain} \cite{perrot2016mapping}, with associated theoretical guarantees \cite{redko2017theoretical}.
Moreover, it is applicable to different transfer scenarios, including unsupervised \cite{courty2017joint} and semi-supervised \cite{yan2018semi} situations. 
Initially, limited by the space complexity of OT (super-quadratically with the size of the sample), it can only be deployed to tackle problems of small or medium size\cite{flamary2021pot}.
Recently, more and more work has attempted to combine deep learning method with OT to train through multiple rounds of minibatch iterative optimization, such as DeepJDOT \cite{damodaran2018deepjdot} and RWOT \cite{xu2020reliable}, breaking the limit of complexity.  
In our setting,  there are few labeled patients in the target hospital, which can be viewed as a problem of semi-supervised transfer learning.  In contrast to the common approach in the unsupervised case, we can further consider the coupling constraints for labeled samples when using OT. 

\textbf{Data Imbalance.}
Traditionally, ML algorithms may assume that the number of samples in considered classes are roughly the same, which is not the case in real-life problems. 
In many medical datasets, the ratio of minority class to majority class can be 1:10, even up to 1:50 \cite{yang2020rethinking}. 
The key point of imbalanced learning is that the minority classes are often more important, 
namely, we need to focus on the diseased samples rather than the healthy samples.
A series of studies have been conducted to overcome data imbalance issue, which can be classified into three types: 
i) data-level methods, which adjust the dataset to balance the minority and majority. A typical way is to downsample the majority or oversample the minority; ii) algorithm-level methods, which do not change the dataset directly, but rather enhance the attention of the model to the minority by modifying the algorithm, i.e., by setting a cost matrix in cost-sensitive learning with help from domain experts\cite{elkan2001foundations};
iii) the combination of both, which takes the advantage of the above methods.
For instance, SMOTEBoost\cite{chawla2003smoteboost} combines SMOTE\cite{chawla2002smote} with Boosting\cite{freund1996experiments} ensemble learning to gain a strong ensemble classifier, SPE \cite{liu2020self} tries to handle tasks on the highly imbalanced, noisy and large-scale dataset by introducing the “classification hardness” function and undersampling with an iterated strategy.

\textbf{Self-paced Learning.} It is a learning paradigm to generates the sequence of training samples by the learner itself, whose core idea is to adaptively select the most informative samples in each iteration \cite{DBLP:conf/nips/KumarPK10}.
In recent years, the self-paced learning regime has been adopted for various tasks, including weakly supervised object detection \cite{DBLP:journals/ijcv/ZhangHZM19}, co-saliency detection \cite{DBLP:journals/pami/ZhangMH17}, and data imbalance \cite{liu2020self}, which indicates the effectiveness of such a learning paradigm.
Though there are some studies that have combined self-paced learning with deep learning for joint learning \cite{DBLP:journals/pami/ZhangHYX20}, we try to combine self-paced learning with optimal transport to eliminate the impact of data imbalance on semi-supervised domain adaptation.

\section{Preliminaries}
In this section, we first define our research problem from an application perspective. Then, we abstract the problem in a transfer learning setting.

\subsection{Sepsis Early Detection}
The objective is to use patients' demographic and physiological data for Sepsis early detection.
Considering the early warning of Sepsis is potentially life-saving, we will detect sepsis 6 hours before the clinical diagnosis of Sepsis. This setting is consistent with 
 the PhysioNet Computing in Cardiology Challenge 2019 \cite{reyna2019early} \cite{goldberger2000physiobank}, whose topic is \textit{Early Prediction of Sepsis from Clinical Data}.\footnote{https://physionet.org/content/challenge-2019/1.0.0/}

In short, given a set of $n$ patients' clinical variables since they entered the ICUs, 
$\{\mathcal{X}_{1}, \mathcal{X}_{2},\cdots, \mathcal{X}_{n}\}$, where the $i$-th patient's is $\mathcal{X}_{i}=\langle \boldsymbol{x}_{1}, \boldsymbol{x}_{2}, \cdots, \boldsymbol{x}_{m}\rangle$, $\boldsymbol{x}_{j}$ is the clinical features of $j$-th time windows (we set the length of one time window as 6 hours). 
Then we aim to predict whether Sepsis will occur in the next 6 hours for each $\boldsymbol{x}_{j}$. 
Thus, it can be seen as a binary classification problem. The clinical variables will be explained in detail in Sec. \ref{sec:feature}.

\subsection{Semi-supervised Transfer Learning Formulation}
When we try to build the detection model in a target hospital with few labeled data, the basic idea is to learn knowledge from other rich data sources. 
In other words,  we can consider this problem as semi-supervised transfer learning.   

In particular, we are given a source domain and a target domain with the same features. 
The source domain contains a large number of labeled samples, and the target domain only contains a limited number of labeled samples (i.e. most samples are unlabeled). 
The task is to improve the classification accuracy in the target domain.
We denote the source domain as $\mathcal{D}^{s} = \{(\boldsymbol{x}_{i}^{s}, y_{i}^{s}) | i=1, 2, \cdots, n_{s}\}$, $\boldsymbol{x}_{i}^{s} \in \mathbb{R}^{d_{s}}$, the target domain as $\mathcal{D}^{t} = \{\mathcal{D}^{l}, \mathcal{D}^{u}\}$ where the labeled data $\mathcal{D}^{l} = \{(\boldsymbol{x}_{j}^{l}, y_{j}^{l}) | j =1, 2, \cdots, n_{l}\}$ and the unlabeled data $ \mathcal{D}^{u} = \{(\boldsymbol{x}_{k}^{u})|k=1,2,\cdots, n_{u}\}$, $\boldsymbol{x}_{j}^{l}, \boldsymbol{x}_{k}^{u} \in \mathbb{R}^{d_{t}}$.
$n_{l}$ and $n_{u}$ are the number of labeled and unlabeled target samples, respectively, $n_{t} = n_{l} + n_{u} \ (n_{l} \ll n_{u})$.
We suppose $d_{s} = d_{t}$ (the source and target domains share the same features) and $\mathcal{Y}^{s} = \mathcal{Y}^{t} = \{0, 1\}$ (binary classification task, 1/0 indicates that Sepsis would/wouldn't happen in 6 hours).

\section{Methodology}
\label{sec:SPSSOT}
In this section, we propose a semi-supervised transfer learning framework, \method, to address our research problem, whose schematic diagram is illustrated as Fig.~\ref{fig:overall}. 
It consists of three main parts: (1) Feature Engineering, (2) Semi-supervised Optimal Transport, and (3) Self-paced ensemble.

\subsection{Feature Engineering}
\label{sec:feature}
We extract the clinical variables and Sepsis criteria from the Electronic Medical Records (EMRs).
For each patient, 34 clinical variables are constructed, including 7 vital sign variables, 23 laboratory variables , and 4 demographic variables. Detailed variables are listed in Table~\ref{table:All features used in model}. The Sepsis-3 criteria are extracted as suspected infection with associated organ dysfunction (SOFA $\geq$ 2) \cite{10.1001/jama.2016.0287} \cite{10.1001/jama.2016.0288}. 

In particular, prior work has shown that typical vital signs, such as heart rate (HR), oxygen saturation ($O_2Sat$), body temperature (Temp), mean arterial blood pressure (MAP), and respiratory rate (Resp), would impact the Sepsis early detection over time \cite{reyna2019early} \cite{yang2020explainable}. Besides, Sepsis incidence rate also varies with respect to the ICULOS (i.e., time stay in ICU). At the early phase, the incidence rate is moderate and slightly increases probably due to the patient prior conditions; at the middle phase, the incidence rate drops a little and becomes stable; at the late phase, the incidence rate increases drastically because there is big vulnerability for the patients that stay long in the ICU\cite{li2020time}.

To capture the time series fluctuation, we take 6 hours as a time window.
In the time slot, we calculate the maximum values, minimum values, means, standard deviations and number of non-missing for all vital signs and laboratory values, while keeping the latest values. 
Finally we concatenate these statistics with demographic variables as the final features to predict whether Sepsis will occur in the next 6 hours.

\begin{table}[]
\caption{Feature Description}
\label{table:All features used in model}
\centering
\begin{tabular}{ll}
\toprule
\textbf{Measurement} & \textbf{Description}\\
\midrule
\multicolumn{2}{l}{\textit{\textbf{Vital sign variables}}} \\
HR        & Heart rates (beats per minute) \\   
$O_2Sat$  & Pulse oximetry (\%)  \\   
Temp      & Temperature ($^{\circ} C$)  \\   
SBP       & Systolic BP (mm Hg)  \\   
MAP       & Mean arterial pressure (mm Hg)  \\   
DBP       & Diastolic BP (mm Hg) \\   
Resp      & Respiration rate (breaths per minute)  \\
\midrule
\multicolumn{2}{l}{\textit{\textbf{Laboratory variables}}}\\
BaseExcess   & Excess bicarbonate (mmol/L)   \\   
$HCO_3$      & Bicarbonate (mmol/L)    \\   
$FiO_2$      & Fraction of inspired oxygen (\%)      \\   
pH           & pH value \\   
$PaCO_2$     & \makecell[l]{The partial pressure of carbon dioxide from \\ arterial blood (mm Hg)} \\  
$SaO_2$      & Oxygen saturation from arterial blood (\%)   \\   
AST          & Aspartate transaminase (IU/L)    \\   
BUN          & Blood urea nitrogen (mg/dL)  \\   
Alkalinephos    & Alkaline phosphatase (IU/L)  \\   
Calcium      & Calcium (mg/dL)   \\   
Chloride    & Chloride (mmol/L)  \\   
Creatinine    & Creatinine (mg/dL)  \\   
Bilirubin direct & Direct bilirubin (mg/dL) \\   
Glucose          & Serum glucose (mg/dL)    \\   
Lactate          & Lactic acid (mg/dL)    \\   
Magnesium       & Magnesium (mmol/dL)       \\   
Phosphate       & Phosphate (mg/dL)      \\   
Potassium   & Potassiam (mmol/L)        \\   
Hct    & Hematocrit (\%)    \\   
Hgb  & Hemoglobin (g/dL)  \\   
PTT     & Partial thromboplastin time (seconds) \\   
WBC     & Leukocyte count (count/L)  \\   
Platelets    & Platelet count (count/mL) \\
\midrule
\multicolumn{2}{l}{\textit{\textbf{Demographic variables}}} \\
Age        & Age (years)     \\   
Sex        & Female (0) or male (1)   \\   
HospAdmTime & Hours from hospitalization to ICU admission\\   
ICULOS       & \makecell[l]{Length of stay in ICU (hours since admission to ICU) } \\   
\bottomrule      
\end{tabular}
\end{table}

\begin{figure*} 
    \centering
    \subfigure[The Architectures of \textit{SSOT}: (1) Initialize the feature generator $\mathcal{G}$ and the classifier $\mathcal{F}$; \quad \quad (2) Fix $\mathcal{G}$ and $\mathcal{F}$, find the current best coupling $\hat{\gamma}$ between $\mathcal{D}^{s}$ and $\mathcal{D}^{t}(\{\mathcal{D}^{l}, \mathcal{D}^{u}\})$ by the OT solver, then fix $\hat{\gamma}$ and update the parameters of $\mathcal{G}$ and $\mathcal{F}$; (3) Iterative training.]
    {\includegraphics[width=0.6\textwidth]{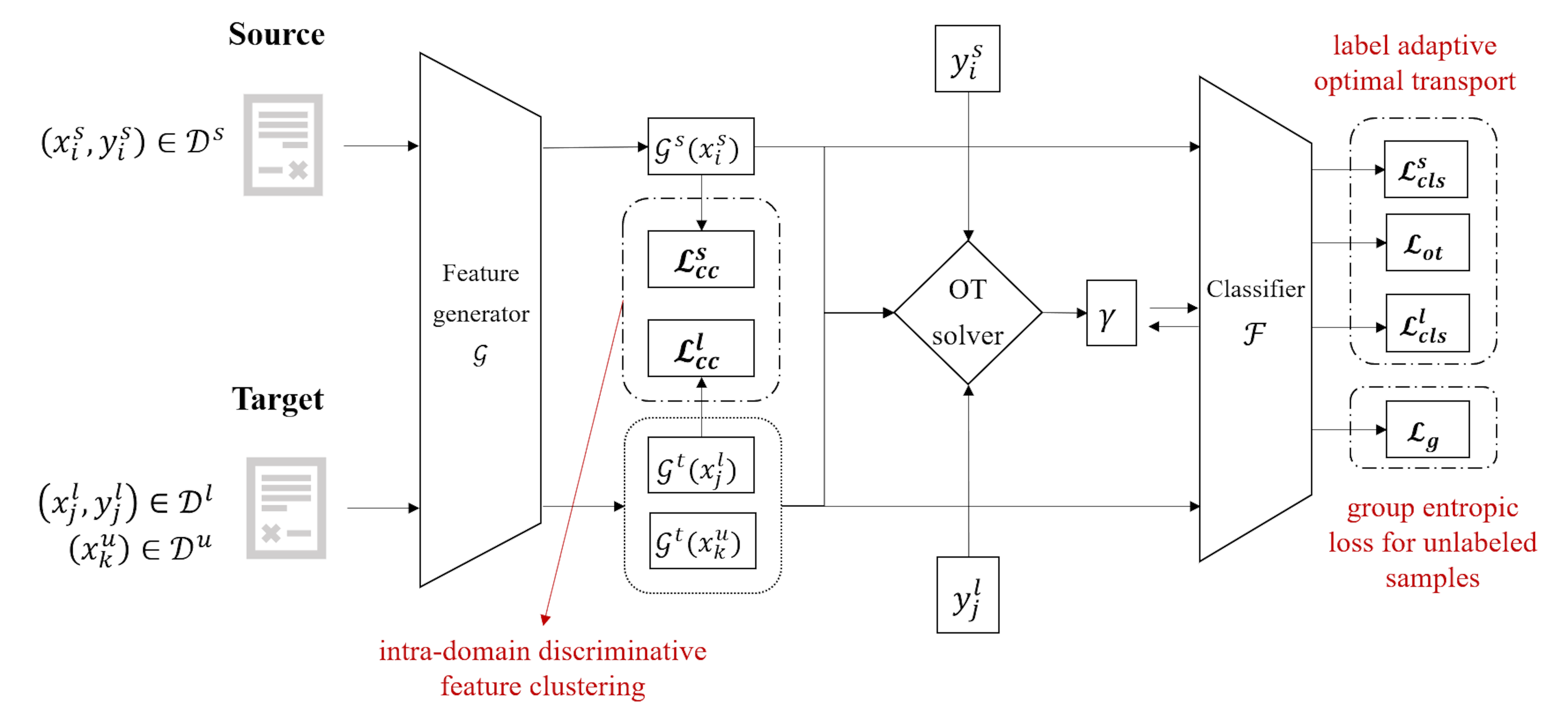}}
    \subfigure[The Objectives of SSOT: (1) the marginal distributions of two domains are identical; (2) the samples belonging to the same class are more aggregated.]
    {\includegraphics[width=0.35\textwidth]{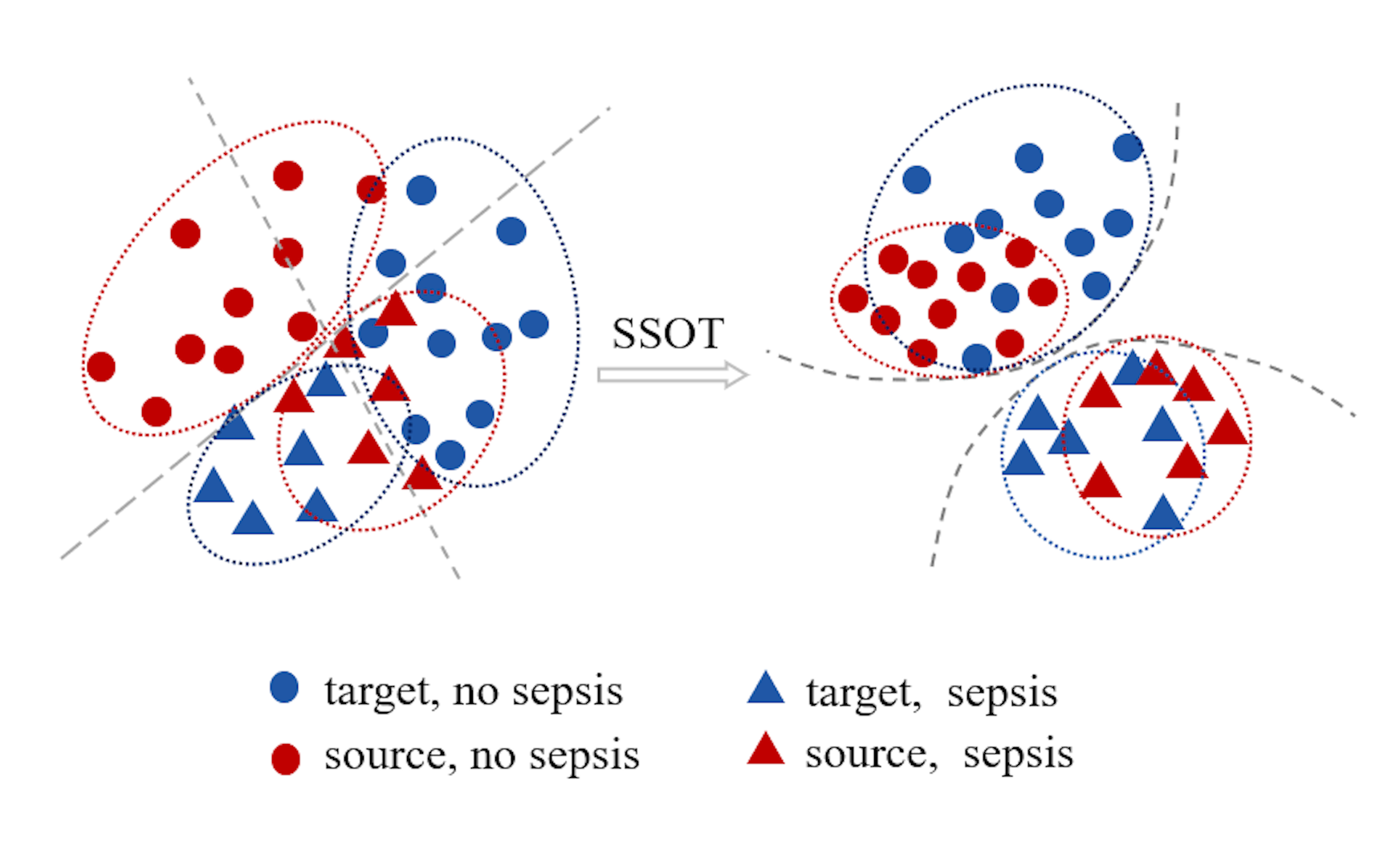}}
    \caption{Semi-supervised Domain Adaptation with Optimal Transport (\textit{SSOT}).}
    \label{fig:otda}
\end{figure*}

\subsection{Semi-supervised Optimal Transport}
\label{sec:SSOT}
Optimal transport theory is a promising strategy applied in transfer learning research. Most existing studies leverage it in unsupervised transfer learning \cite{courty2016optimal}. That is, they assume that no labeled data is in the target domain. 
However, few labeled samples in the target domain are more in line with the real situation \cite{9578015}. For instance, it is usually acceptable to label a few samples when we want to deploy a Sepsis early detection system in a new hospital, if this can significantly improve the system performance. 

With this in mind, we design a novel semi-supervised optimal transport (\textit{SSOT}) strategy for transfer learning. The purpose is to increase classification performance by minimizing the distribution discrepancy between the source and target domains with a well-structured neural network. Specifically, the neural network enables the end-to-end training of a transferable feature generator and an adaptive classifier, as illustrated in Fig.~\ref{fig:otda}. 
Because clinical features can be viewed as tabular data, we choose classical multi-layer perception (MLP) with shared weights as the feature generator $\mathcal{G}$.

The main parts of \textit{SSOT} are \textit{label adaptive optimal transport}, \textit{group entropic loss for unlabeled samples} and \textit{intra-domain discriminative feature clustering}. In the following sections, we will present the details of \textit{SSOT}.

\subsubsection{Label Adaptive Optimal Transport}
In the traditional optimal transport strategy for unsupervised transfer learning, the source and target samples are mapped to a shared feature space where the samples of both domains cannot be differentiated \cite{courty2016optimal}.
In semi-supervised transfer learning, there are \textit{a few labeled samples} in the target domain;
we then need to propose a new optimal transport strategy to effectively leverage these labeled target data. 
Therefore, beyond the traditional optimal transport strategy, our mechanism further conducts the optimization to ensure that \textit{the labeled target samples should only be matched with the same-labeled source samples.}

\textbf{Optimal Transport}. The optimization of optimal transport is based on Kantorovich problem \cite{angenent2003minimizing} which seeks for a general coupling $\gamma \in \mathcal{X}(\mathcal{D}^{s}, \mathcal{D}^{t})$ between $\mathcal{D}^{s}$ and $\mathcal{D}^{t}$:

\begin{equation} \small
\label{eq:ot}
   \gamma^{*}=\mathop{\arg \min}\limits_{\gamma \in \mathcal{X} (\mathcal{D}^{s}, \mathcal{D}^{t})} \int_{\mathcal{D}^{s} \times \mathcal{D}^{t}} \mathcal{C}({x}^{s}, {x}^{t}) d \gamma({x}^{s}, {x}^{t})
\end{equation}
where $\mathcal{X} (\mathcal{D}^{s}, \mathcal{D}^{t})$ denotes the probability distribution between $\mathcal{D}^{s}$ and $\mathcal{D}^{t}$. 

The discrete reformulation is 
\begin{equation} \small
    \gamma^{*} = \mathop{\arg \min}\limits_{\gamma \in \mathcal{X} (\mathcal{D}^{s}, \mathcal{D}^{t})} \langle \gamma, \mathcal{C} \rangle_{F}
\end{equation}
where $\langle\cdot, \cdot\rangle_{F}$ is the Frobenius dot product, $\mathcal{C} \in \mathbb{R}^{n_{s} \times n_{t}}$ is the cost function matrix. $\mathcal{C}(x^{s}, x^{t}) = \|x^{s} - x^{t} \|^{k}$ represents the cost to move probability mass from $x^{s}$ to $x^{t}$; we set $k=2$ following the literature \cite{courty2016optimal}.

\textbf{Label Adaptive Constraint}. As a few data can be labeled in the target domain, we adjust the cost of transport according to  the labels of the two domains' samples. 
If two samples have the same labels, it means that the transport cost is very low between these two samples. Therefore, we can use a parameter, $\rho$, to adjust the cost, i.e, $\mathcal{C}(x^{s}, x^{t}) = \rho$, if $y(x^{s}) = y(x^{t})$; otherwise, we set the cost to $1$.
At the same time, for unlabeled target samples, we can consider supplementing a weight for transport cost to measure the difference between the predicted probabilities and the labels of source samples. 
Accordingly, we design a reweight matrix, called \textit{label adaptive matrix} $\mathcal{R}$. 
Then, the label adaptive optimal transport can be written as 
\begin{equation} \small
\label{eq:ot2}
    \gamma^{*} = \mathop{\arg \min}\limits_{\gamma \in \mathcal{X} (\mathcal{D}^{s}, \mathcal{D}^{t})} \langle\gamma, \mathcal{R} \cdot \mathcal{C}\rangle_{F}
\end{equation}
where  
\begin{equation} \footnotesize
    \label{eq:label_matrix}
    \mathcal{R}(x^{t},x^{s}) = \begin{cases}
    \rho + (1-\rho) \cdot |y(x^{s}) - y(x^{t})| \in \{\rho, 1\}, (x^{t}, y^{t}) \in \mathcal{D}^{l} \\
    \rho + (1-\rho) \cdot |y(x^{s}) - \hat{y}(x^{t})| \in [\rho, 1], x^{t} \in \mathcal{D}^{u} \nonumber
    \end{cases} ,
\end{equation}
$y(x)$ is the label of a sample $x$ and $\hat y(x)$ is the prediction probability $P(y(x)=1)$.

In summary, the solution to this problem can be described to minimize the following objective function
\begin{equation} \small
\label{eq:loss_lot}
    \mathcal{L}_{lot} = \alpha \mathcal{L}_{ot} + \theta_{s} \mathcal{L}_{cls}^{s} + \mathcal{L}_{cls}^{l}
\end{equation}
where $\alpha$ and $\theta_{s}$ are hyper-parameters, $\mathcal{L}_{ot}$ is the cost of optimal transport, $\mathcal{L}_{cls}^{s}$ and $\mathcal{L}_{cls}^{l}$ are the cross entropy function of the source and target domain, i.e.,

\begin{equation} \small
    \label{eq:loss_ot}
    \mathcal{L}_{ot} = \sum \limits_{i, j} \gamma^{*}_{i,j} (\|\mathcal{G}(x_{i}^{s}) - \mathcal{G}(x_{j}^{t})\|^{2})
\end{equation}

\begin{equation} \small
\label{eq:loss_cls}
\begin{aligned}
    \mathcal{L}_{cls}^{s} = - \sum \limits_{x_{i}^{s} \in X_{s}} y(x_{i}^{s})\log \hat{y}(x_{i}^{s}) \\
    \mathcal{L}_{cls}^{l} =  -\sum \limits_{x_{j}^{l} \in X_{l}} y({x_{j}^{l}}) \log \hat{y}(x_{j}^{l})
\end{aligned}
\end{equation}

\subsubsection{Group Entropic Loss for Unlabeled Samples}
It is insufficient to ensure that the mappings of source and target samples cannot be differentiated in a shared feature space.
This only implies that the marginal distributions of the two domains are identical.
To further mitigate the differences in the conditional distributions, we borrow the labels of the source domain to calculate the classification loss of target unlabeled samples. 
That is, if one target sample $x_{j}^{u}$ has a high transport probability from one source sample $x_{i}^{s}$ ($\gamma_{ij}$ is high), then it is probable that the predicted label is the same with  $y(x_{i}^{s})$. 

Based on this idea, we form the group entropic loss to compare the cross entropy between the predicted probability of each target unlabeled sample and the true label of each source sample. 
It can be written as
\begin{equation} \small
\label{eq:loss_group}
\begin{aligned}
    \mathcal{L}_{g} = -\frac{1}{n_{s}} \frac{1}{n_{u}} \sum_{x_{i}^{s} \in X^{s}} \sum_{x_{j}^{t} \in X^{u}} \gamma^{*}_{ij} (y(x_{i}^{s})\log \hat{y}(x_{j}^{u}))
\end{aligned}
\end{equation}
where $\hat{y}(x_{j}^{u}) = \mathcal{F}(\mathcal{G}(x_{j}^{u}))$ is the predicted probability of $x_{j}^{u}$. 
By penalizing couplings with high cross entropies, we can achieve that each unlabeled target sample can be transported from source samples with same class.

\subsubsection{Intra-domain Feature Discrimination}
The center loss is originally proposed to enhance the discriminative power of the deeply learned features for face recognition  \cite{wen2016discriminative}.
Inspired by this, we also hope to ensure that samples belonging to the same class are close to each other in the feature space.
Here, we consider the discriminative centroid loss $\mathcal{L}_{cc}$ for the labeled samples in both source domain and target domain.

\begin{equation} \small
\label{eq:loss_cc}
\begin{aligned}
    \mathcal{L}_{cc} = & \frac{1}{n_{s}}\sum_{i=1}^{n_{s}}\|\mathcal{G}(x_{i}^{s}) - c_{i}^{s}\|^{2}_{2} - (\|c_{0}^{s} - c_{1}^{s}\|_{2}^{2}) \\
    & + \frac{1}{n_{l}} \sum_{j=1}^{n_{l}}\|\mathcal{G}(x_{j}^{t}) - c_{j}^{t}\|^{2}_{2}  -  (\|c_{0}^{t} - c_{1}^{t}\|_{2}^{2})
\end{aligned}
\end{equation}

\begin{algorithm}[htp]
\footnotesize
\caption{Semi-supervised Optimal Transport (\textit{SSOT})}
\label{alg:SSOT} 
\begin{algorithmic}[1] 
\Require 
Source data as $\mathcal{D}^{s} = \{(\boldsymbol{x}_{i}^{s}, y_{i}^{s})\}_{i=1}^{n_{s}}$; Target labeled data as $\mathcal{D}^{l} = \{(\boldsymbol{x}_{j}^{l}, y_{j}^{l})\}_{j=1}^{n_{l}}$; Target unlabeled data as $\mathcal{D}^{u} = \{(\boldsymbol{x}_{k}^{u})\}_{k=1}^{n_{u}}$; \textit{T} is set as the total number of training iterations; \textit{n} represents the batch-size for training.
\State Initialize the feature generator $\mathcal{G}$ and the classifier $\mathcal{F}$ by fine tuning; 
\label{code:fram:initialize} 
\For { $i = 1$ to \textit{T} }
\label{code:fram:loop} 
\State Randomly select half of source samples and target labeled samples;
\State Calculate the class centers in two domains according to Eq.\ref{eq:center_source}. 
\State Randomly choose source samples $\{(x_{i}^{s}, y_{i}^{s})\}_{i=1}^{n} \in \mathcal{D}^{s}$, target labeled samples $\{(x_{j}^{l}, y_{j}^{l})\}_{j=1}^{n/2} \in \mathcal{D}^{l}$, and target unlabeled samples $\{(x_{k}^{u})\}_{k=1}^{n/2} \in \mathcal{D}^{u}$;
\State Fix $\hat{\mathcal{G}}$ and $\hat{\mathcal{F}}$, solve for $\gamma$;
\label{code:fram:update_gamma}
\State Fix $\hat{\gamma}$, update parameters of $\mathcal{G}$ and  $\mathcal{F}$;
\label{code:fram:update_classifier}
\EndFor \\ 
\Return $\mathcal{G}$ and $\mathcal{F}$; 
\end{algorithmic} 
\end{algorithm}

\noindent where $c_{i}^{s}$ and $c_{j}^{t}$ denote the corresponding class center of $x_{i}^{s}$ and $x_{j}^{t}$ in the source domain and target domain, respectively. 
We evaluate them by averaging the deep discriminative features of the samples in the corresponding class.

\begin{equation} \small
\label{eq:center_source}
\begin{aligned}
    &c_{k}^{s} = \frac{1}{S_{a}} \sum_{i=1}^{N_{a}}\mathcal{G}(x_{i}^{s}) \mathcal{I}(y_{i}^{s} , k) , k \in \{0,1\} \\
    &c_{k}^{t} = \frac{1}{S_{b}} \sum_{j=1}^{N_{b}}\mathcal{G}(x_{j}^{t}) \mathcal{I}(y_{j}^{t} , k)
    , k \in \{0,1\}
\end{aligned}
\end{equation}
where $\mathcal{I}(y_{i} , k) = \begin{cases} 1,& y_{i}=k \\ 0,& y_{i} = 1-k \end{cases}$, and $S_{a} = \sum_{i=1}^{N_{a}} \mathcal{I}(y_{i}^{s}, k)$, $S_{b} = \sum_{j=1}^{N_{b}} \mathcal{I}(y_{j}^{b}, k)$. Ideally, the class centers should be calculated based on all the samples while the procedure is time-consuming. Herein, we compute the class centers by randomly sampling $N_{a}$ and $N_{b}$ samples. In our experiments, we set $N_{a}=\frac{1}{2}\times n_{s}, N_{b}=\frac{1}{2} \times n_{l}$.

\subsubsection{Training}
Here, we introduce the training process of \textit{SSOT}. 
Considering the three parts of \textit{SSOT}, the training objective can be described as

\begin{equation} \small
\label{eq:loss_sum}
    \min \limits_{\mathcal{G}, \mathcal{F}} \mathcal{L}_{lot} + \lambda \mathcal{L}_{g} + \beta \mathcal{L}_{cc}
\end{equation}
where $\lambda$ and $\beta$ denote hyper-parameters that trade-off the contribution of the intra-domain structures and domain alignment, respectively. 

The training process is shown in Algorithm \ref{alg:SSOT}. 
Specifically, in each iteration, we use two steps to update the parameters. First, we fix the feature generator $\mathcal{G}$ and the classifier $\mathcal{F}$, and use the optimal transport mechanism (POT (python optimal transport) \cite{flamary2021pot} in our implementation) to calculate the coupling $\gamma$ (line \ref{code:fram:update_gamma}); Second, we fix $\gamma$ to update $\mathcal{G}$ and $\mathcal{F}$ with the stochastic gradient descent algorithm (line \ref{code:fram:update_classifier}).

\subsection{Self-paced Ensemble}
When we try to transfer knowledge from the source domain to the target domain, label shift may happen between two domains \cite{azizzadenesheli2019regularized}. That is, the label distribution changes from the source to the target. 
At the same time, it is ubiquitous that medical diagnostic data is extremely imbalanced. This reveals not only the disproportion between classes but also other difficulties embedded in the nature of data, such as \textit{noises and class overlapping} \cite{gamberger1999experiments}. 

\begin{figure}
    \centering
    \includegraphics[width=1.0\linewidth]{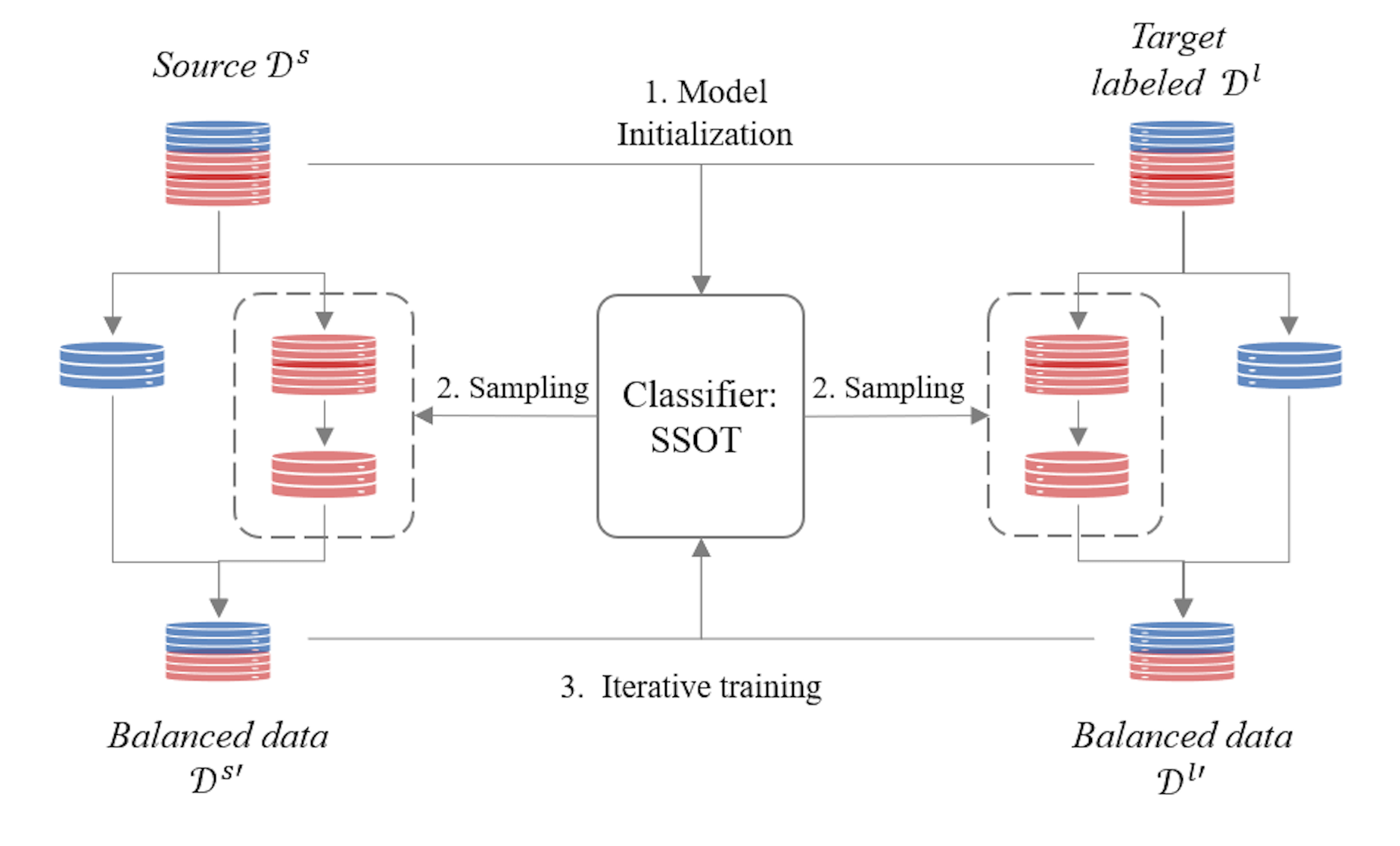}
    \caption{The Core Idea of Self-paced Ensemble Based on SSOT (SPSSOT). There are 3 main steps: (1) Initialize SSOT according to Algorithm. \ref{alg:SSOT}; (2) Self-paced under-sampling from majority class in both domains to obtain balanced data; (3) Get an additive model by iteration training. \textit{Best viewed in color.}}
    \label{fig:spe}
\end{figure}

Therefore, it is necessary to design a strategy to solve both problems.
Inspired by the self-paced ensemble method (SPE) \cite{liu2020self}, we adapt it to our end-to-end transfer-learning prediction framework so as to sample labeled data simultaneously from two domains. 
Fig.\ref{fig:spe} shows the core idea of our SPE-enhanced SSOT (SPSSOT) mechanism. 
Next, we will present in detail how to carry out the sampling strategy.

\subsubsection{Classification Hardness Function}
We use $\mathcal{H}$ to denote the hardness function, which can be calculated by the summation of individual sample errors, such as absolute error and cross entropy. Given a classifier $\mathcal{F}$ and a sample $(x, y)$, the hardness can be written as $\mathcal{H}(x, y, \mathcal{F})=|\mathcal{F}(x)-y|\in [0, 1]$. This value contains information that is highly associated with the difficulty of the classification, like noise and model capacity. According to the hardness values, we can divide samples into three types as follows:
\begin{itemize}
    \item \textit{Trivial samples} account for the largest proportion of samples that are easy to classify, i.e., each of them only contributes tiny hardness. However, the overall contribution is not negligible due to the large number of samples.
    \item \textit{Noise samples} are different from trivial samples. Though the sample number is small, each sample has a large hardness value. These samples can be caused by the indistinguishable overlapping and will exist stably even when the model is converged.
    \item \textit {Borderline samples} are the rest training samples. 
\end{itemize}

Intuitively, while sampling, we should (1) keep a small proportion of trivial samples to maintain the original data distribution to avoid overfitting; (2) exclude the interference of noise samples during training; (3) enlarge the weights of borderline samples to improve the model performance. What remains to be settled is how to distinguish three types of samples and achieve under-sampling in practice.

\subsubsection{Self-paced Under-sampling}
There are two important components, \textit{self-paced hardness harmonize} and \textit{combination with SSOT} in achieving self-paced sampling and iterative training. 
Algorithm \ref{alg:spe} describes the detailed process.

\textbf{Self-paced hardness harmonize.} We regard the class with a higher proportion as the \textit{majority} (in Sepsis early detection, the majority class is the patients without Sepsis).
After calculating the hardness values of the majority samples, we can split them into $k$ bins regarding different hardness levels, i.e., the $l$-th bin, $B_{l}$, is defined as
\begin{equation} \small
\label{eq:Bin_l}
    B_l = \{(x, y) | \frac{l-1}{k} \leq \mathcal{H}(x, y, \mathcal{F}) <  \frac{l}{k}\}, \mathcal{H} \in [0, 1]
\end{equation}

\begin{algorithm}[htp]
\footnotesize
\caption{Semi-supervised Optimal Transport with Self-paced Ensemble (\textit{SPSSOT})}
\label{alg:spe} 
\begin{algorithmic}[1] 
\Require Source data as $\mathcal{D}^{s} = \{(\boldsymbol{x}_{i}^{s}, y_{i}^{s})\}_{i=1}^{n_{s}}$; Target labeled data as $\mathcal{D}^{l} = \{(\boldsymbol{x}_{j}^{l}, y_{j}^{l})\}_{j=1}^{n_{l}}$; Target unlabeled data as $\mathcal{D}^{u} = \{(\boldsymbol{x}_{k}^{u})\}_{k=1}^{n_{u}}$;Hardness function $\mathcal{H}$; Base classifier \textit{SSOT}; Number of base classifiers $n$; Number of bins $k$; Total number of training iterations of \textit{SSOT} \textit{T}; 
\State Initialize $\textit{SSOT}_{0}$ according to Algorithm \ref{alg:SSOT}; 
\For {$i = 1$ to $n$}
\State Ensemble {\footnotesize $F_{i}(\mathcal{D}^{s}, \mathcal{D}^{l}, \mathcal{D}^{u}) = \frac{1}{i} \sum_{j=0}^{i-1} SSOT_{j}(\mathcal{D}^{s}, \mathcal{D}^{l}, \mathcal{D}^{u})$}; 
\For {$\mathcal{D} \in \{\mathcal{D}^{s}, \mathcal{D}^{l}\}$}
\State Initialize $\mathcal{P} \Leftarrow$ minority in $\mathcal{D}$;
\State Cut majority set into $k$ bins w.r.t. $\mathcal{H}(\mathcal{D}, F_{i})$: $B_{1}, B_{2}, \cdots, B_{k}$;
\State Average hardness contribution in $l$-th bin: $h_{l} = \sum_{m\in B_{l}} \mathcal{H}(x_{m}, y_{m},  F_{i}) / |B_{l}|, {\forall} l = 1, \cdots, k$;
\State Update self-paced factor $\omega = tan(\frac{i\pi}{2n})$;
\label{code:fram:self-paced} 
\State Unnormalized sampling weight of $l$-th bin: {\small$p_{l} = \frac{1}{h_{l}+\omega}$}, ${\forall} l=1,\cdots,k$;
\State Under-sample from $l$-th bin with $\frac{p_{l}}{\sum_{m}p_{m}}\cdot |\mathcal{P}|$;
\EndFor
\State Train $SSOT_{i}$ using newly under-sampled subset according to Algorithm \ref{alg:SSOT};
\EndFor
\\
\Return Final ensemble model {\small $F(\mathcal{D}^{s}, \mathcal{D}^{l}, \mathcal{D}^{u}) = \frac{1}{n}\sum_{m=1}^{n}$ $\textit{SSOT}_{m}(\mathcal{D}^{s}, \mathcal{D}^{l}, \mathcal{D}^{u})$}; 
\end{algorithmic} 
\end{algorithm}

\noindent Then we can under-sample from every bin by ensuring that the total hardness contribution of each bin is the same, so as to generate a balanced dataset. 
By harmonizing hardness contribution, the sampling probability of those bins with a larger population will be generally lower. 
Moreover, we leverage a self-paced factor $\omega$ to adjust the decreasing level in training process. 
This factor is calculated by $tan$ function (line \ref{code:fram:self-paced} of Algorithm \ref{alg:spe}) \cite{liu2020self}.
As $\omega$ gets larger, the sampling weights of hard samples will increase. 
In the beginning, we pay more attention to borderline samples to improve model performance.
While in the later iterations ($\omega$ becomes very large), the model still keeps a certain number of trivial samples as the “skeleton” to avoid overfitting.

\textbf{Combination With \textit{SSOT}.} Unlike the original SPE \cite{liu2020self} that is applied on supervised classification models (e.g., classification models of sklearn\footnote{https://scikit-learn.org/stable/}), we should solve the data imbalance in two domains and combine the self-paced ensemble strategy with \textit{SSOT}. 
As shown in Fig.\ref{fig:spe}, we perform self-paced under-sampling simultaneously on source labeled data and target labeled data; then, we obtain two balanced datasets (both source and target domains) for training SSOT iteratively.

\section{Experiments}
\subsection{Dataset}
We conduct our experiments on two widely-used real-life Sepsis detection  datasets, \textbf{MIMIC-III} \cite{johnson2016mimic} and the PhysioNet Computing in Cardiology Challenge 2019 \cite{reyna2019early} (\textbf{Challenge}). 
Specifically, we extracted the first 48-hour data since patients entered ICUs. As a part of patients' records have a large number of missing values, we screened out the patients whose missing value ratio is less than 80\%. 
To obtain the dynamic change information of the data over a period of time, for every patient, we calculated the maximum, minimum, mean, standard error and latest of each clinical indicator within 6 hours.
In this way, a patient's 48-hour ICU stay can be converted to eight 6-hour records (samples). Then, we can use the $k$-th ($k \in [1, 8]$) record to predict whether Sepsis would occur or not in the next 6 hours. 
Through such preprocessing, we obtain the final data for experiments. 
Some basic statistic information is enumerated in Table \ref{table:statistic}. 

\begin{table}[t]
\centering
\caption{Statistics of the Datasets}
\label{table:statistic}
\begin{tabular}[t]{lcc}
\toprule
        &\textbf{MIMIC}          &\textbf{Challenge}\\
\midrule
\# patients         & 12529         & 8270    \\
\# septic patients  & 2977          & 1831    \\
Sepsis prevalence (\%)      & 23.76       & 22.14  \\
\midrule
\# samples        & 87501         & 45674    \\
\# samples occur Sepsis in next 6 hours & 5032    & 4869  \\
samples with sepsis (\%) & 5.75    &  10.66 \\
\bottomrule
\end{tabular}
\end{table}

\begin{table*}[]
\centering
\caption{Overall Evaluation Results}
\label{table:Overall_result}
\setlength{\tabcolsep}{6mm}{
\begin{tabular}{lccccc}
\toprule
\multicolumn{1}{c}{} & \multicolumn{2}{c}{\textbf{MIMIC $\to$ Challenge}} & \multicolumn{2}{c}{\textbf{Challenge $\to$ MIMIC}} & Average\\  
\cmidrule(r){2-3}  \cmidrule(r){4-5}
% \noalign{\smallskip} 
\multicolumn{1}{c}{}     & AUC   & improvement & AUC   & improvement  & improvement\\   \hline
\multicolumn{5}{l}{{\emph {\textbf{Source Only}}}}                       \\
\textit{LR}              & 56.15 $\pm$ 0.85 & 15.94\%     & 72.70 $\pm$ 0.35 & 4.61\%   & 10.28\%   \\
\textit{NN}              & 59.24 $\pm$ 0.75 & 9.89\%      & 70.53 $\pm$ 1.34 & 7.83\%   & 8.86\%  \\
\textit{XGBoost}         & 60.81 $\pm$ 0.26 & 7.05\%      & 59.41 $\pm$ 0.94 & 28.01\%  & 17.53\%   \\ \hline
\multicolumn{5}{l}{{\emph {\textbf{Target Only}}}}                      \\
\textit{LR}              & 60.21 $\pm$ 0.07 & 8.12\%      & 71.62 $\pm$ 0.50 & 6.19\%   & 7.16\%    \\
\textit{NN}              & 60.58 $\pm$ 0.14 & 7.46\%      & 61.92 $\pm$ 0.29 & 22.82\%  & 15.14\%   \\
\textit{XGBoost}         & 58.90 $\pm$ 0.65 & 10.53\%     & 72.97 $\pm$ 0.54 & 4.10\%   & 7.38\%    \\ \hline
\multicolumn{5}{l}{{\emph {\textbf{Source \& Target Train Together}}}}                  \\
% \multicolumn{5}{l}{\textbf{Train together}}                         \\
\textit{LR}              & 59.90 $\pm$ 1.15 & 8.68\%      & 72.89 $\pm$ 0.47 & 4.34\%    & 6.51\%  \\
\textit{NN}              & 60.81 $\pm$ 0.24 & 7.05\%      & 71.53 $\pm$ 0.09 & 6.32\%    & 6.69\%  \\
\textit{XGBoost}         & 60.25 $\pm$ 0.35 & 8.05\%      & 68.71 $\pm$ 0.21 & 10.68\%   & 9.37\%  \\ \hline
\multicolumn{5}{l}{{\emph {\textbf{Source \& Target Transfer}}}}                  \\             
\textit{DeepJDOT}       & 61.17 $\pm$ 0.75 & 6.42\%      & 72.64 $\pm$ 0.39 & 4.69\%     & 5.56\%  \\
\textit{Finetune}       & 60.11 $\pm$ 0.73 & 8.30\%      & 71.62 $\pm$ 1.72 & 6.19\%     & 7.25\% \\ 
\textit{MME}            & 61.49 $\pm$ 0.84 & 5.87\%      & 75.07 $\pm$ 0.70 & 1.31\%     & 3.59\% \\
\textit{LIRR}           & 62.76 $\pm$ 0.95 & 3.73\%      & 75.35 $\pm$ 0.59 & 0.93\%     & 2.33\% \\
\textit{$S^{3}D$}            & 61.87 $\pm$ 0.61 & 5.22\%      & 75.56 $\pm$ 0.37 & 0.65\%     & 2.94\% \\
\method (our) & \textbf{65.10 $\pm$ 0.24} & -           & \textbf{76.05 $\pm$ 0.54} & -     & -      \\ 
\bottomrule
\end{tabular}}
\end{table*}

\subsection{Compared Algorithms}
In our experiments, we split the target data into three parts: 1\% as labeled data (we will change the ratio in Sec. \ref{sec:sensitivity}), 79\% as unlabeled data, and  20\% as test data. 
To compare with our method \method, we implement four types of baselines.
\begin{itemize}
    \item \textit{Source only}: train a classifier only with the source data and directly use it with the target test data.
    \item \textit{Target only}: train a classifier only with the target labeled data (i.e., 1\% of the target data) and use it with the target test data.
    \item \textit{Source \& Target Train Together}: put the source data and the labeled target data together as training data to learn a classifier.
    \item \textit{Source \& Target Transfer}: instead of training together, design specific transfer learning methods to transfer knowledge from the source domain to the target domain.   
\end{itemize}

In the former three types, we all use three classical machine learning algorithms popular in Sepsis early detection, i.e., Logistic Regression (\textit{LR}) \cite{shashikumar2017early}, Neural Network (\textit{NN}) \cite{futoma2017learning} and \textit{XGBoost}\cite{zabihi2019sepsis}.
For the fourth type of baselines, we implement five methods for comparison, including an unsupervised domain adaptation method using transport optimal theory, \textit{DeepJDOT} \cite{damodaran2018deepjdot}, fine-tuned \textit{NN} (\textit{Finetune}), and three start-of-the-art semi-supervised domain adaptation methods, \textit{MME} \cite{saito2019semi}, \textit{LIRR} \cite{9578015} and \textit{$S^{3}D$} \cite{Yoon_2022_WACV}.

\subsection{Experiment Design}
There is a self-paced sampling strategy in \method to solve the class imbalance question. For a fair comparison, we also apply the method, SPE \footnote{https://github.com/ZhiningLiu1998/self-paced-ensemble}, to downsample majority data and train ensemble models when using \textit{LR}, \textit{NN} and \textit{XGBoost} as base classifiers, where the hardness function is set to Squared Error, the number of base classifiers is set to 20 and the number of bins is to 15. \textit{LR} and \textit{XGBoost} are trained with default scikit-learn parameters, and \textit{NN} has four linear layers, whose dimension is (256, 128, 128, 2).

In transfer learning methods, we use two linear layers as the feature generator $\mathcal{G}$ and the dimension is (256, 128). 
The structure of classifier, $\mathcal{F}$, is also two-layer and the dimension is (128, 2), where 2 means binary classification in our task. 
The batch size is set to 128, the parameter optimization algorithm is SGD, and the learning rate is set to 0.001. 
In \textit{Finetune}, we first train $\mathcal{G}$ and  $\mathcal{F}$ with source data and fine-tune them with target labeled data; both parts are trained for 100 epoches. In \textit{DeepJDOT}, we set $\alpha=0.5$, $\lambda_{t}=1.0$, $\lambda_{s}=2.0$ and the number of iterations is 5000. 
In \method, we set $\alpha=0.05$ in Eq. \eqref{eq:loss_lot}, $\theta_{s} = 1.0$ in Eq.\eqref{eq:loss_lot} and  $\beta = 0.15$, $\lambda = 0.5$ in Eq. \eqref{eq:loss_sum}. 
In Algorithm \ref{alg:spe}, the hardness function is Squared Error, the number of base classifiers is set to 5, the number of bins is to 10, and the number of iterations is set to 5000. 
In \textit{MME}, \textit{LIRR} and \textit{$S^{3}D$}, we apply the same network structure of $\mathcal{G}$ and $\mathcal{F}$, and keep the same batch size and learning rate.
We repeat each experiment for 5 times and record the average results.
The parameter sensitivity analysis is conducted later in Sec.\ref{sec:sensitivity}.

Because over 80\% papers about Sepsis prediction reported AUC\cite{fleuren2020machine}, we also pick it as our performance metric.

Our experiment platform is a server with AMD Ryzen 9 3900X 12-Core Processor, 64 GB RAM and GeForce RTX 3090. 
We use Python 3.8 with scikit-learn 0.24, POT 0.7 and tensorflow 2.4 on Ubuntu 20.04 for algorithm implementation. 
Our codes and models can be found on Github\footnote{https://github.com/RuiqingDing/SPSSOT}.

% Ablation Study & Convergence
\begin{figure*} 
\begin{minipage}{0.5\textwidth}
    \subfigure[MIMIC $\to$ Challenge]{
    \includegraphics[width=0.43\linewidth]{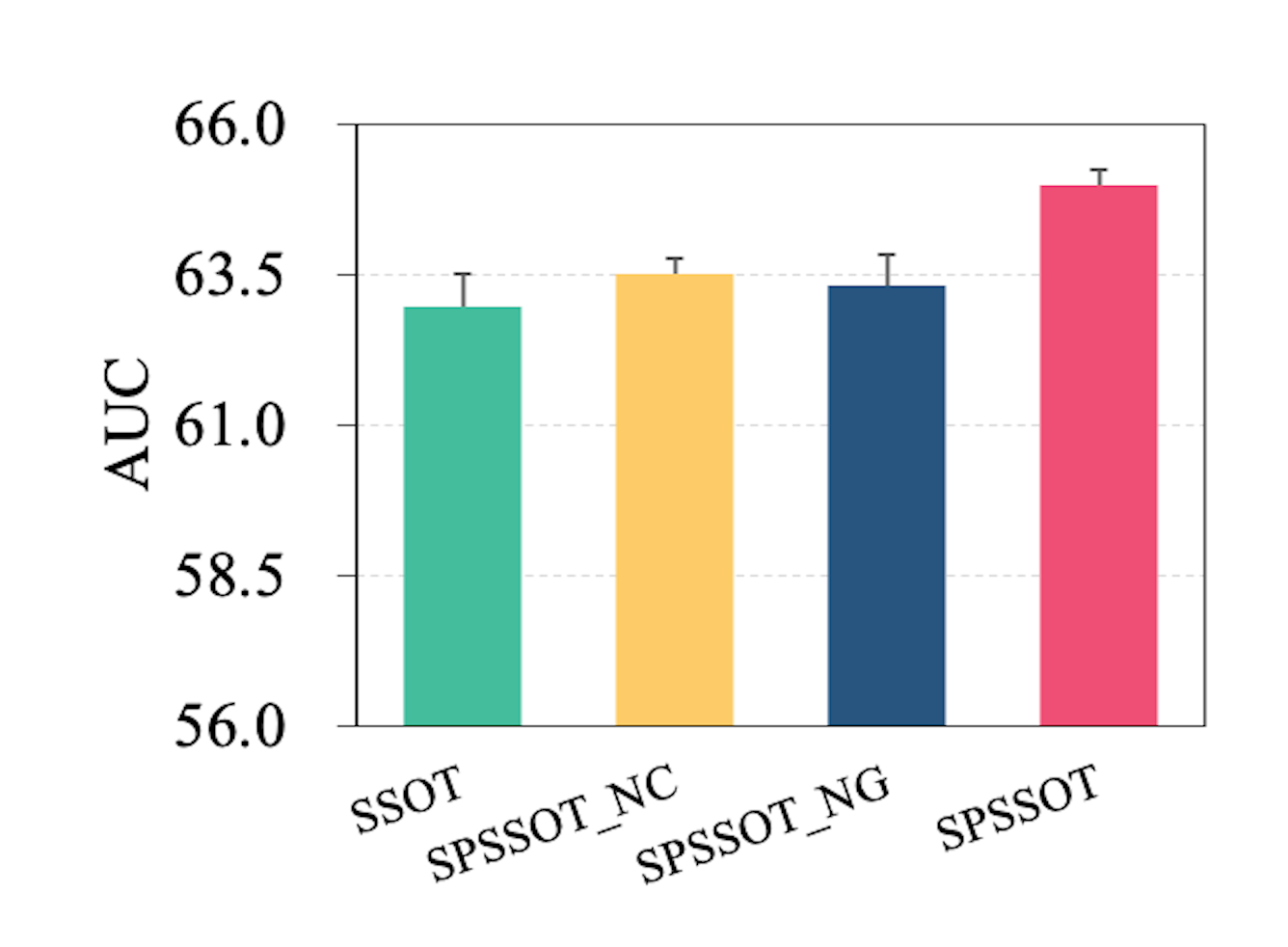}
    }
    \subfigure[Challenge $\to$ MIMIC]{
    \includegraphics[width=0.43\linewidth]{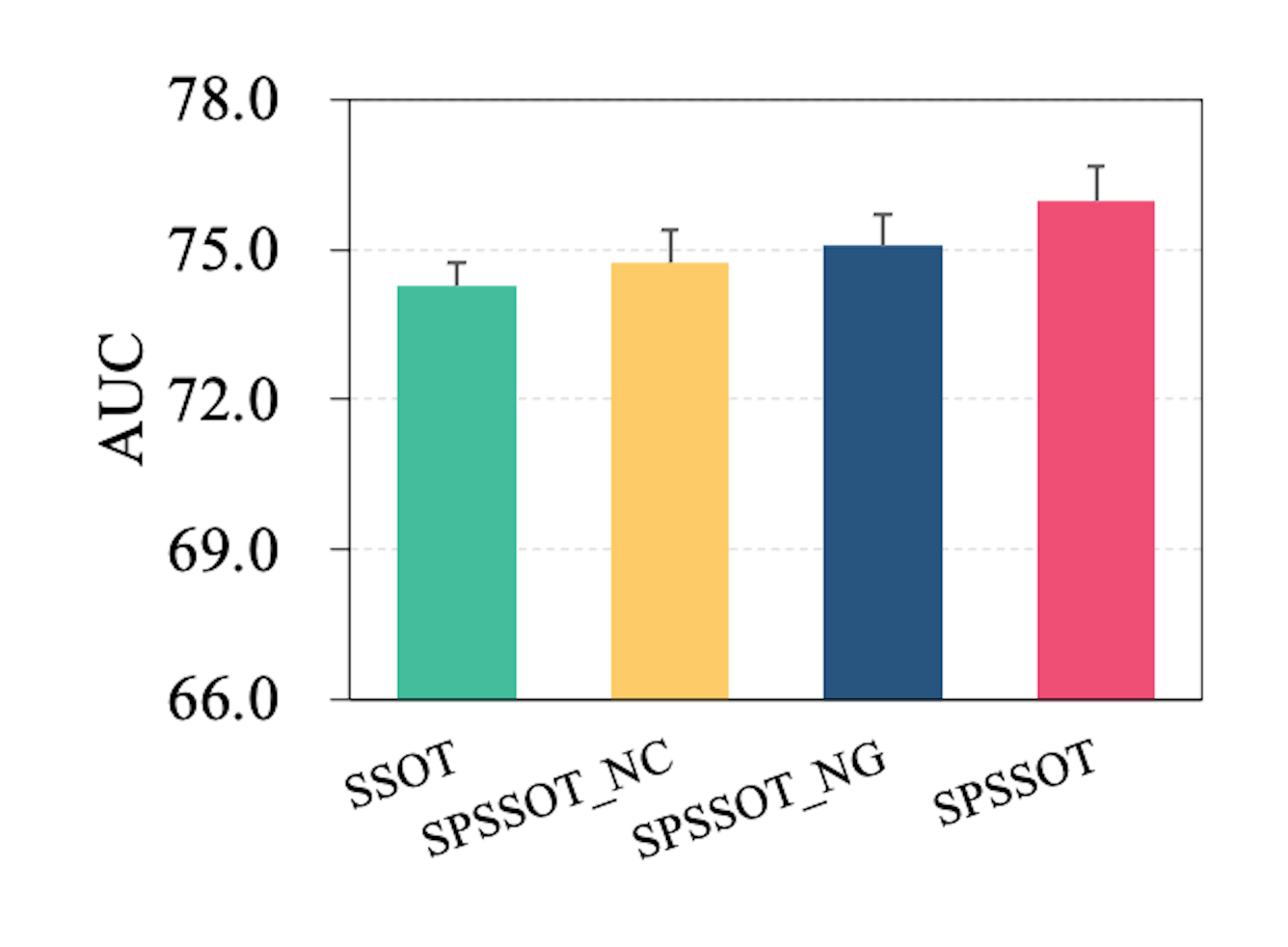}
    }
    \caption{Ablation Study: compare \method with its variants.}
    \label{fig:ablation}
\end{minipage}
\begin{minipage}{0.5\textwidth}
    \centering
    \subfigure[MIMIC $\to$ Challenge]{
    \includegraphics[width=0.43\linewidth]{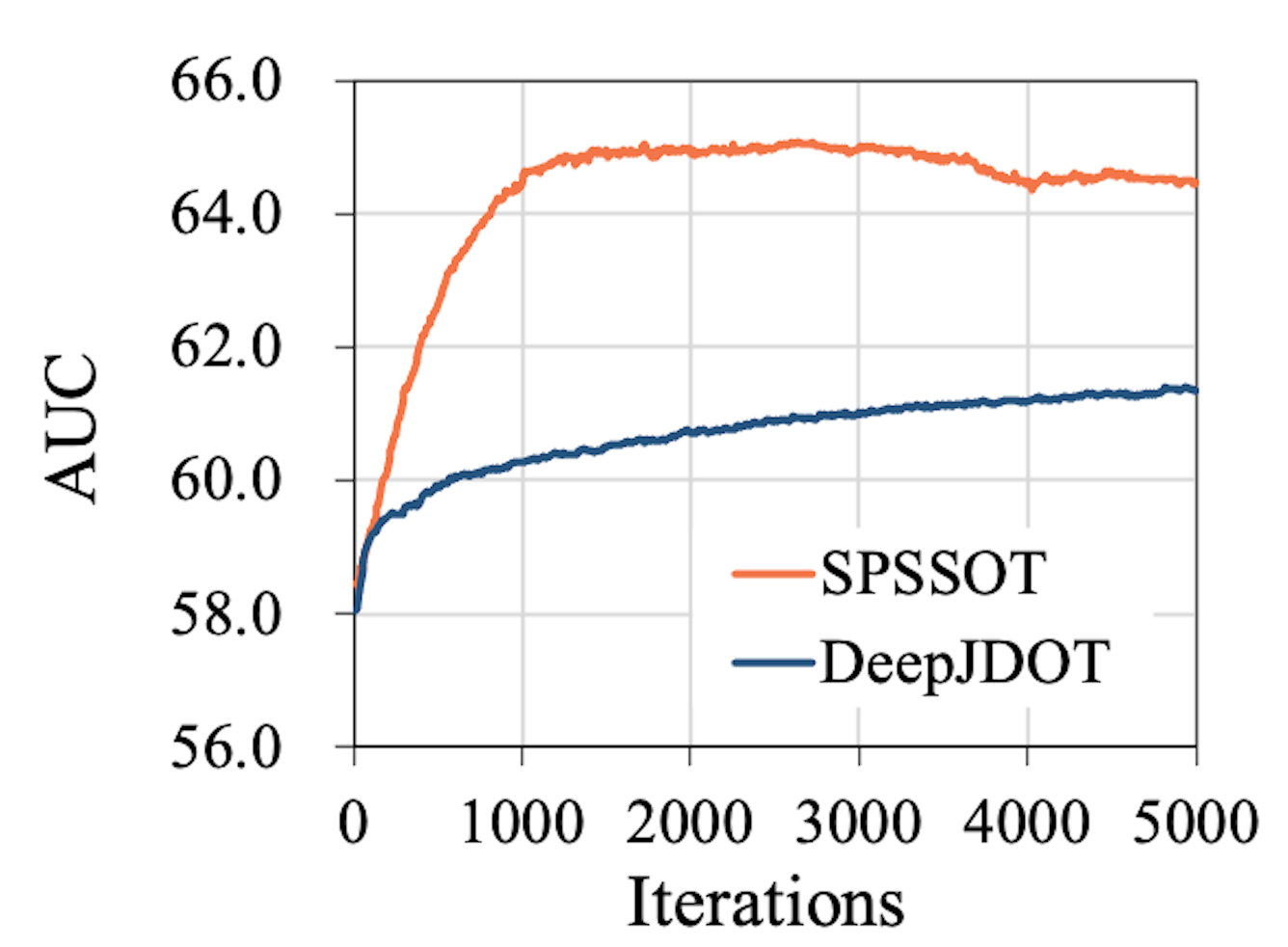}
    }
    \quad
    \subfigure[Challenge $\to$ MIMIC]{
    \includegraphics[width=0.43\linewidth]{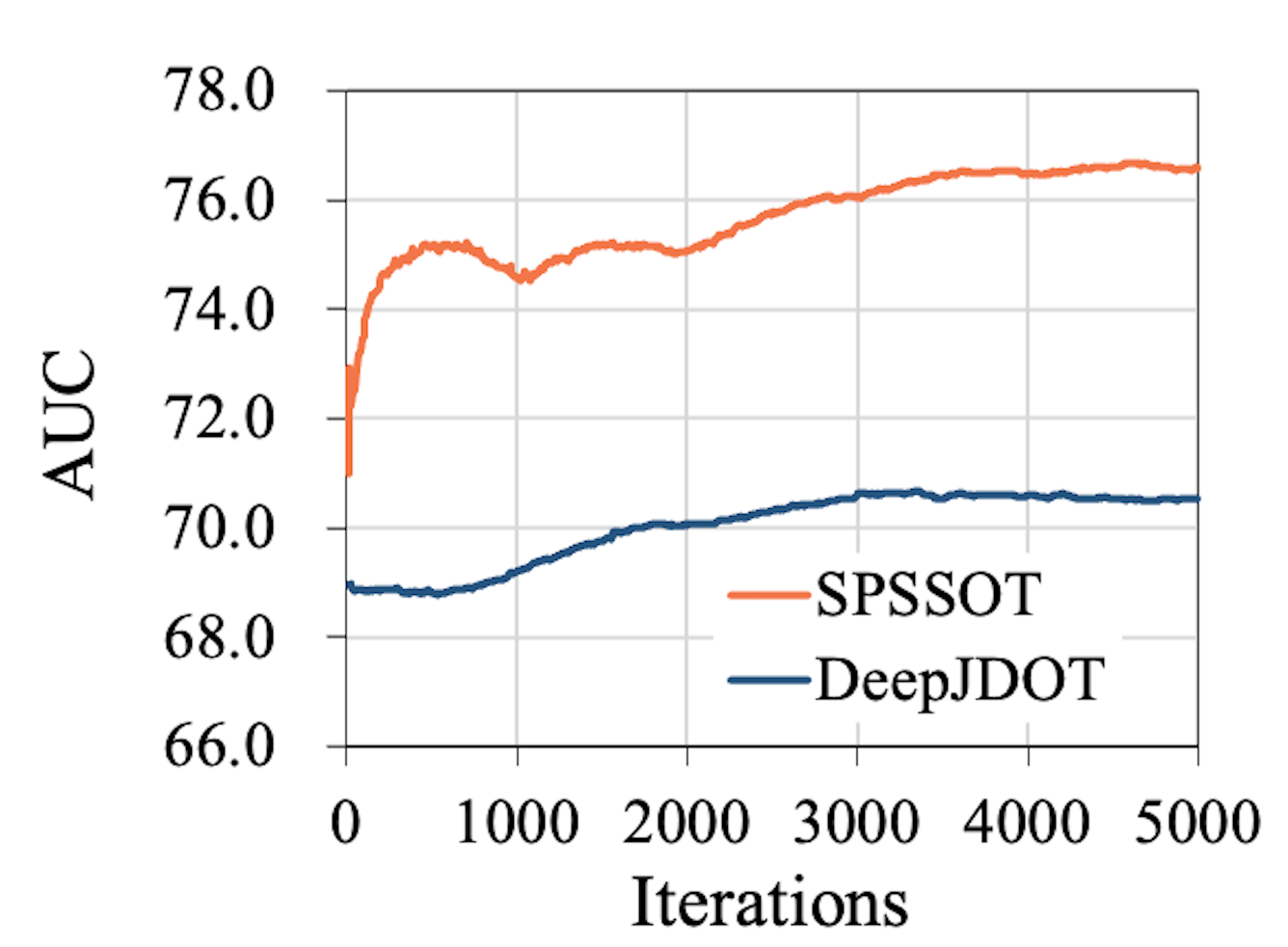}
    }
    \caption{The convergence performance of \method and \textit{DeepJDOT}.}
    \label{fig:convergence}
\end{minipage}
\end{figure*}

% different labeled percentage of target and source
\begin{figure*} 
\begin{minipage}{0.5\textwidth}
    \centering
    \subfigure[MIMIC $\to$ Challenge]{
    \includegraphics[width=0.43\textwidth]{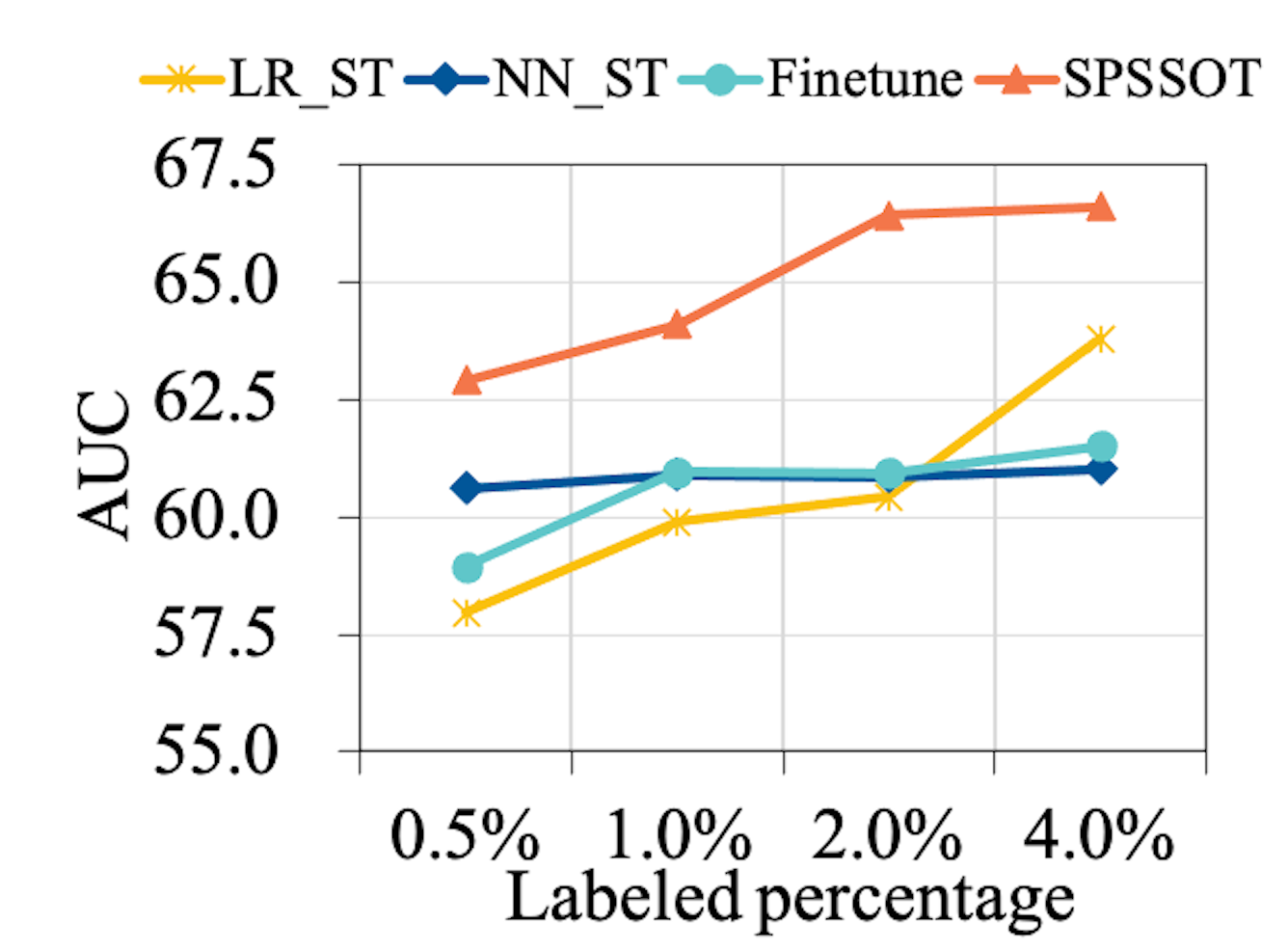}
    }
    \subfigure[Challenge $\to$ MIMIC]{
    \includegraphics[width=0.43\textwidth]{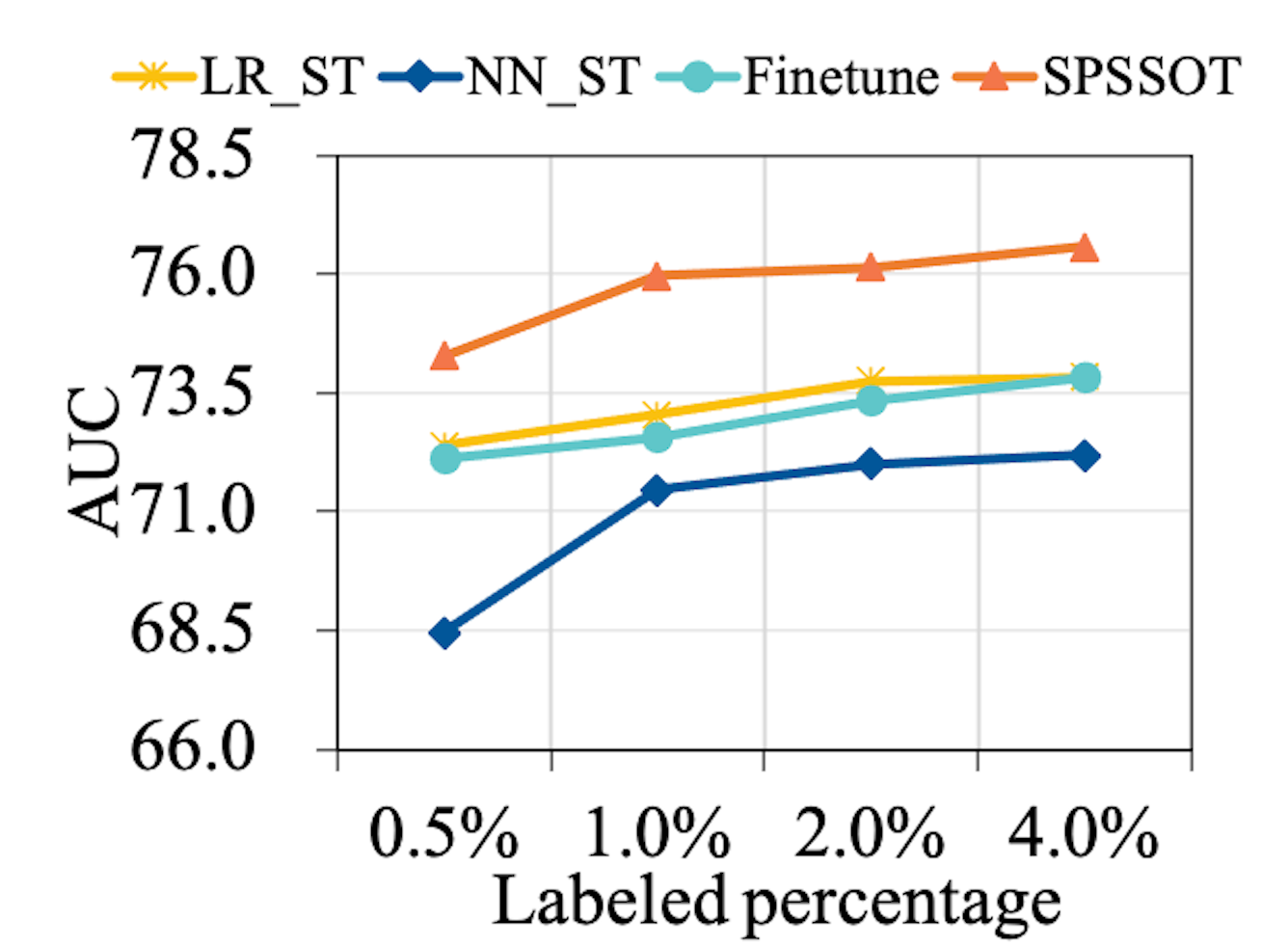}
    }
    \caption{Different percentages of labeled data in target domain.}
    \label{fig:label_percent}
\end{minipage}
\begin{minipage}{0.5\textwidth}
    \centering
    \subfigure[MIMIC $\to$ Challenge]{
    \includegraphics[width=0.43\textwidth]{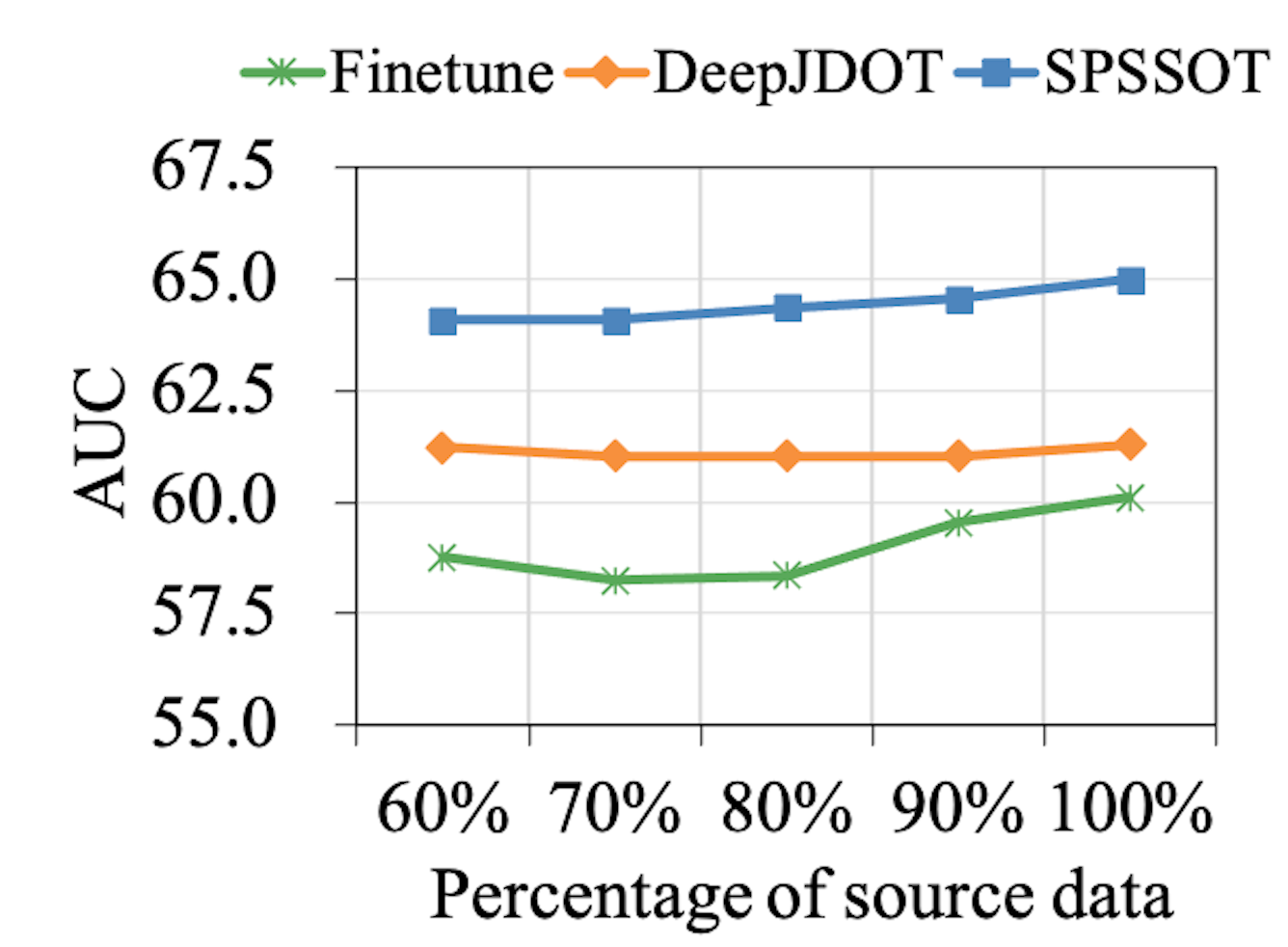}
    }
    \quad
    \subfigure[Challenge $\to$ MIMIC]{
    \includegraphics[width=0.43\textwidth]{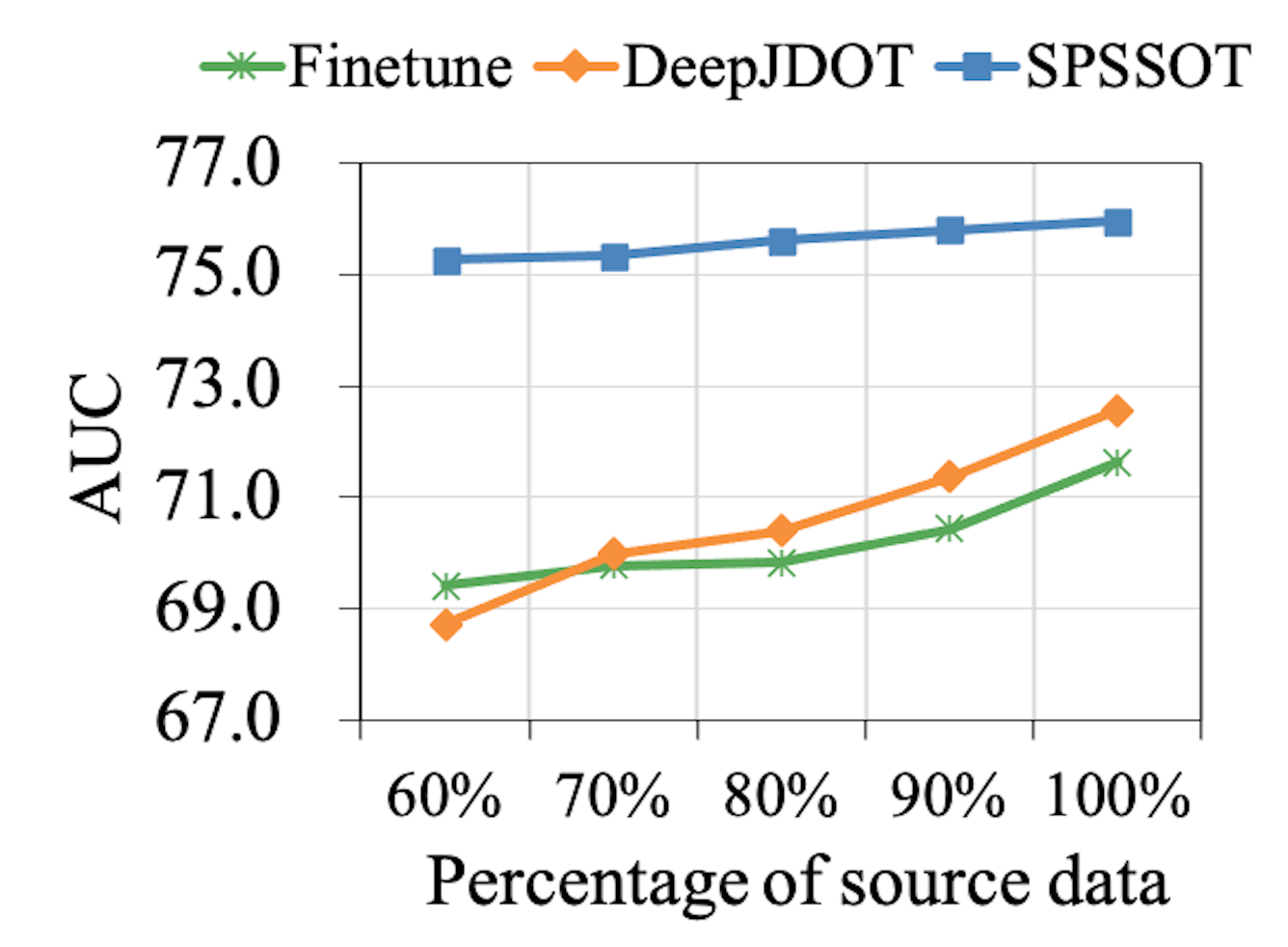}
    }
    \caption{Different sampling percentages of source data.}
    \label{fig:source_percentage}
\end{minipage}
\end{figure*}

% hyper-parameters
\begin{figure*} 
    \centering
    \subfigure[weight of optimal transport]{
    \includegraphics[width=0.21\textwidth]{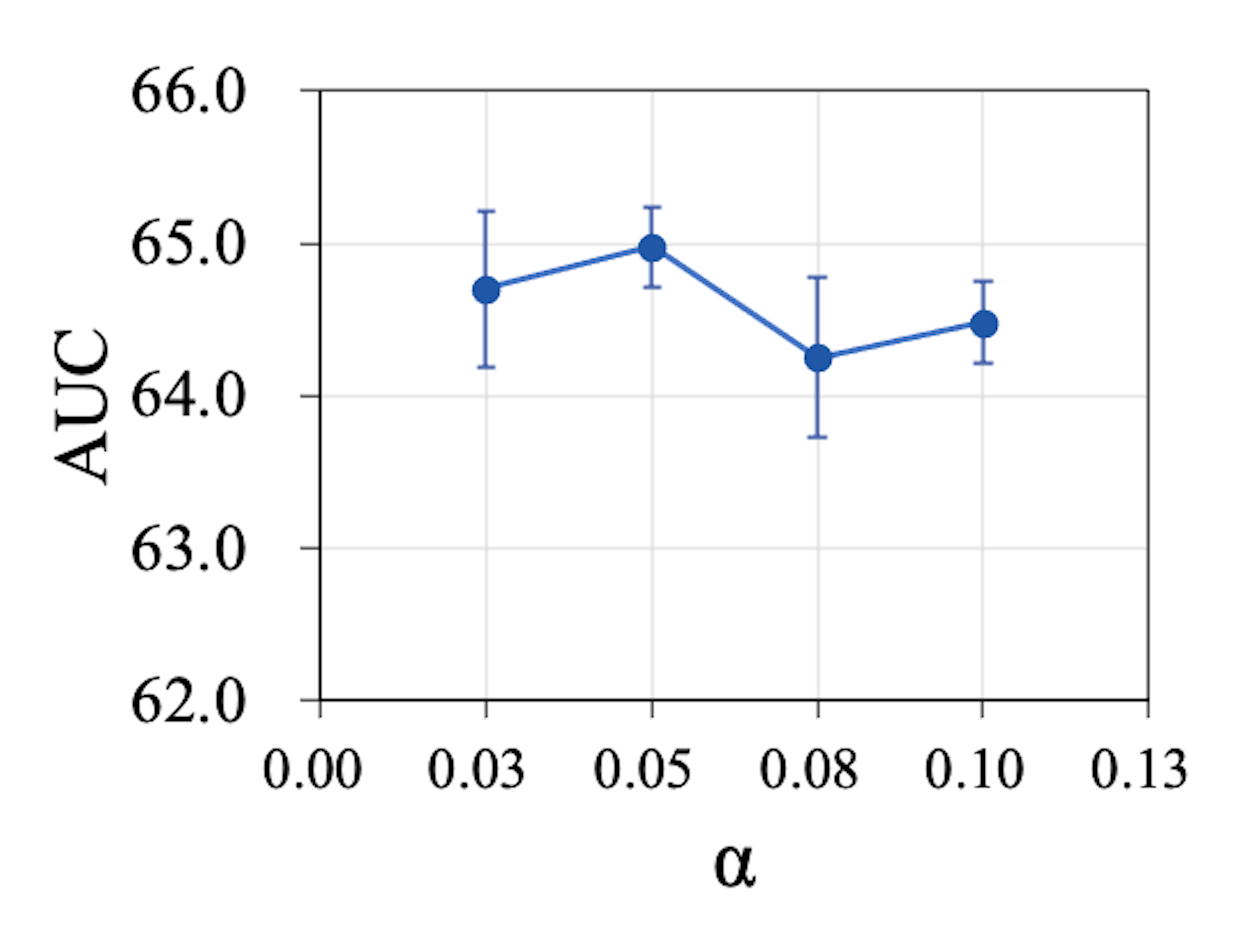}
    }
    \quad
    \subfigure[weight of source's classification loss]{
    \includegraphics[width=0.21\textwidth]{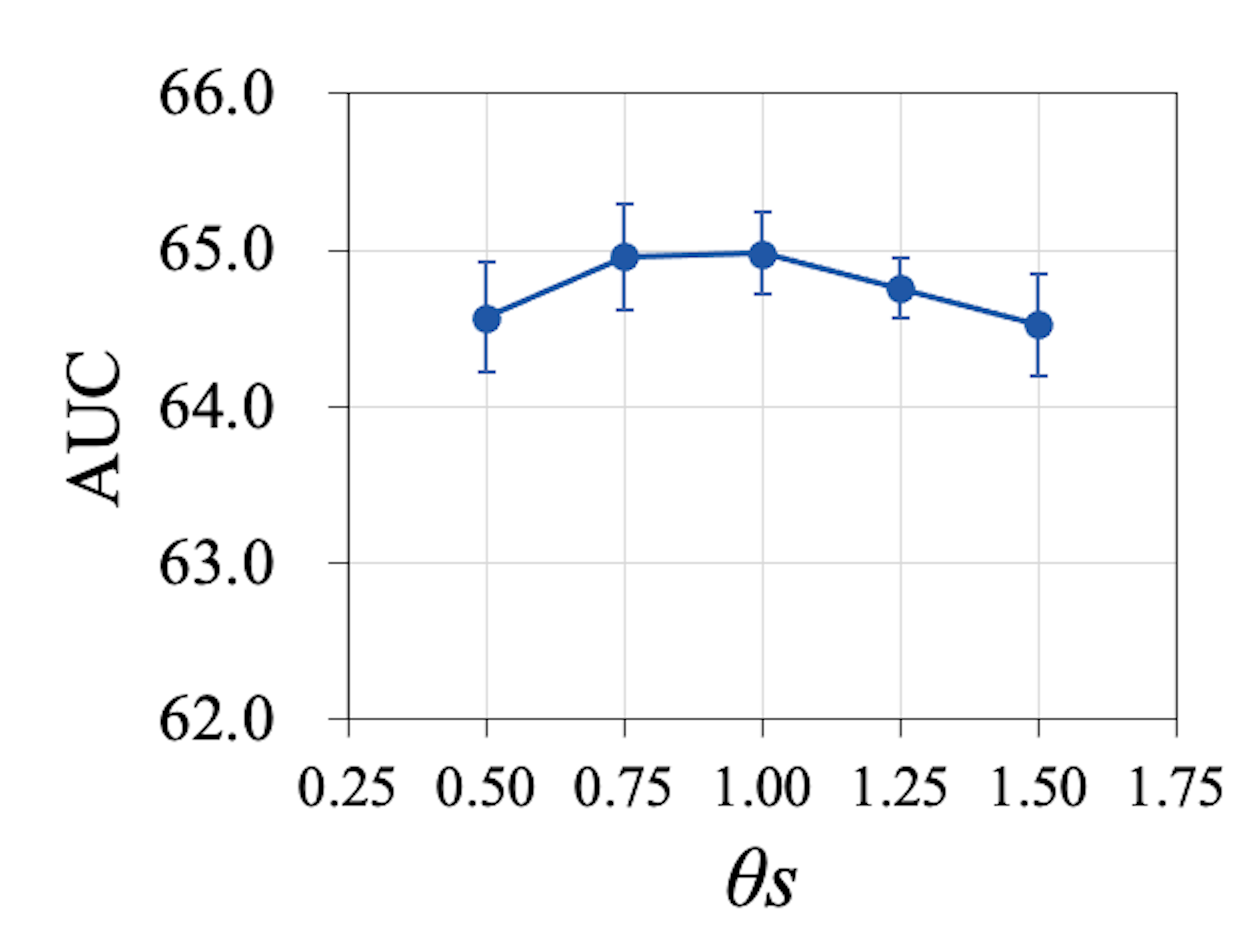}
    }
    \quad
    \subfigure[weight of discriminative centroid loss]{
    \includegraphics[width=0.21\textwidth]{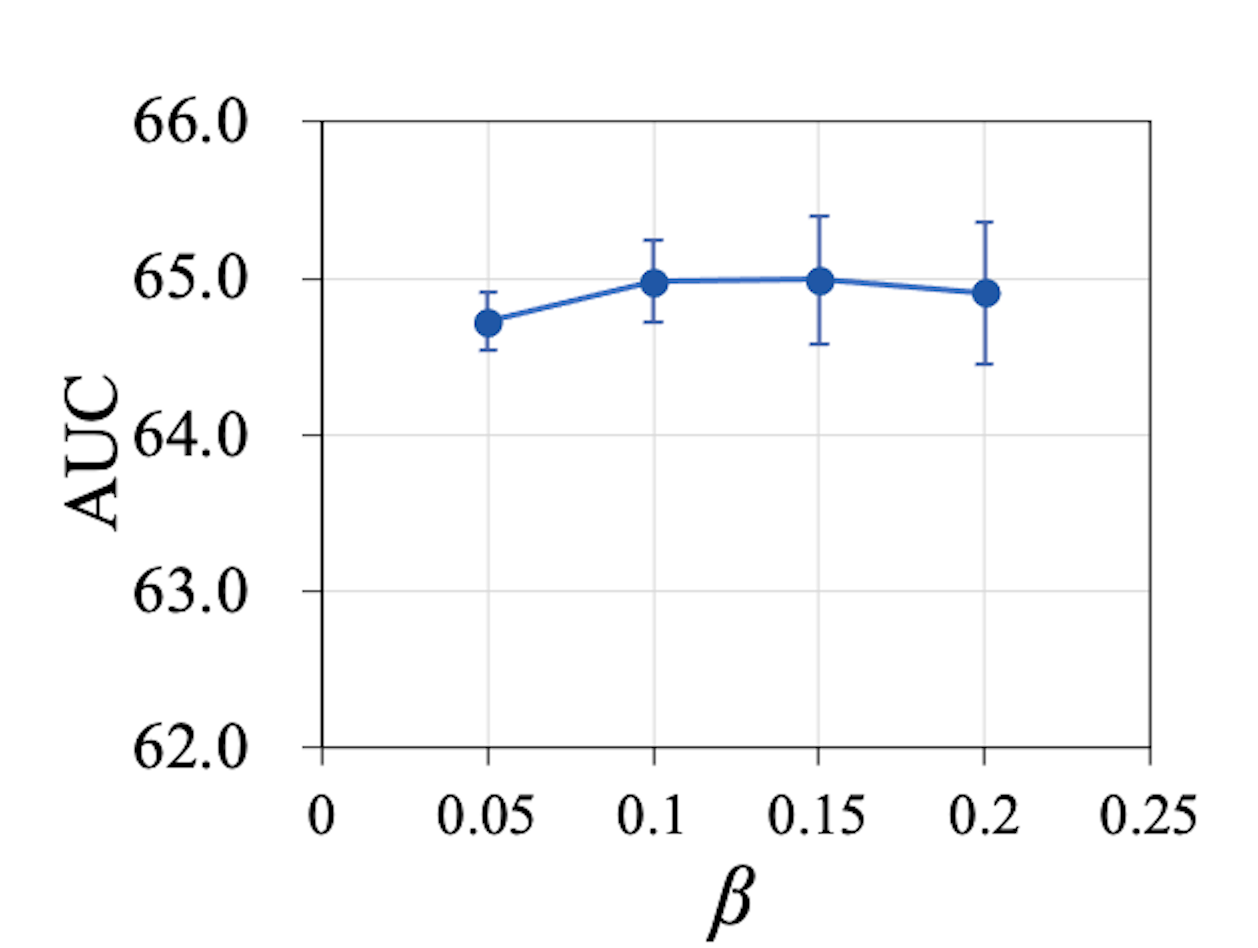}
    }
    \quad
    \subfigure[weight of group entropy]{
    \includegraphics[width=0.21\textwidth]{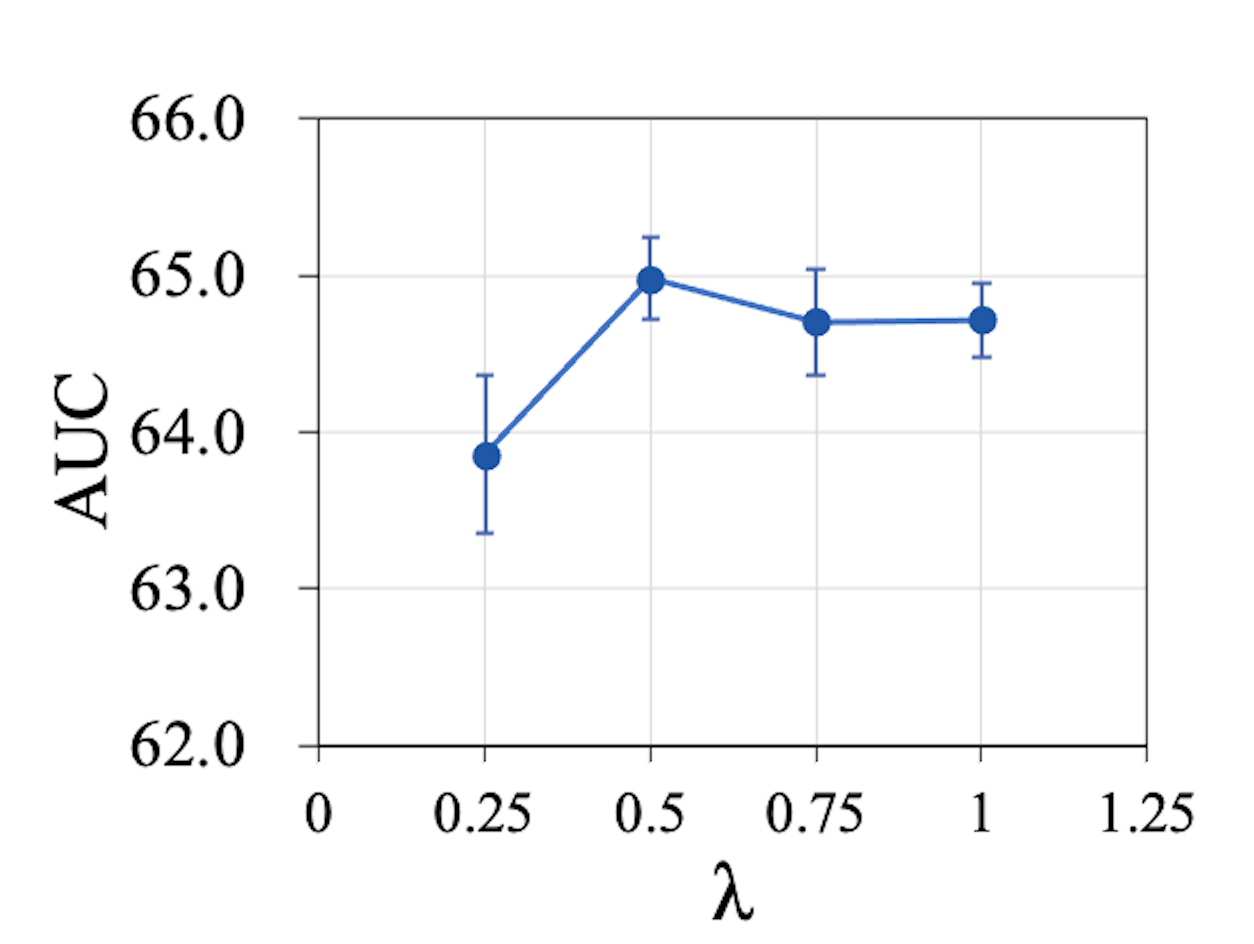}
    }
    \caption{Parameter Sensitivity of MIMIC $\to$ Challenge: vary the four hyperparameters in the loss function and compare the results of the experiments.}
    \label{fig:sensitivity}
\end{figure*}

\subsection{Results and Discussion}
The experiment results of \method and the baselines are reported in Table \ref{table:Overall_result}.
To make a more comprehensive comparison, we demonstrate the experimental results from three perspectives.

First, \method outperforms the other five transfer learning baselines. 
Between them, \textit{DeepJDOT} is an unsupervised transfer learning method based on Optimal Transport \cite{damodaran2018deepjdot}. Thus, the improvement is expected because our \method can further leverage the target labeled data (although the labeled data may be little).
Compared to \textit{Finetune}, a common method in transfer learning, the advantage of our method further verifies the effectiveness of using optimal transport to align two feature spaces during the training process. 
It is worth noting that, while \textit{Finetune} considers 1\% labeled data in the target domain, its performance is even worse than \textit{DeepJDOT} without considering any labeled target data. 
This indicates that even if we have certain labeled data in the target domain, it is still non-trivial to properly leverage the knowledge of such labeled data. 
In addition, \textit{MME}, \textit{LIRR} and \textit{$S^{3}D$} are the state-of-the-art semi-supervised transfer learning methods. They all outperform the other baselines, which shows that it makes sense to use the labeled data of the source and target domains for domain adaptation at the same time. 
Specifically, these methods are comparable to \method in Challenge $\to$ MIMIC, while \method is over 3\% ahead in MIMIC $\to$ Challenge. 
To some degree, this demonstrates that our method can more efficiently use the sparsely labeled target data throughout the knowledge transfer process.

Second, compared to the baseline methods (\textit{LR, NN} and \textit{XGBoost}) that are trained with source data and target labeled data together, \method improves at least 7.05\% in MIMIC $\to$ Challenge and 4.34\% in Challenge $\to$ MIMIC. 
The probable reason is that the feature distributions of two domains are different so that simply putting two domains' data together for training is not effective. 
When we further compare the models that \textit{train together} and \textit{Finetune}, \textit{Finetune} may not perform better. 
This result illustrates that though the model trained with source data provides initial parameters for \textit{FineTune}, the initialization probably is not suitable for target data. 
Therefore, it may appear a negative transfer when the feature distributions of two domains are not similar.

Third, all the no-transfer baselines, i.e., \textit{Source Only} and \textit{Target Only}, perform rather poorly. 
The results of \textit{Source Only} indicate that, though MIMIC and Challenge both are medical datasets with the same features, there still are some differences. 
For \textit{Target Only} methods, as we only have a small amount of labeled data (1\% in our setting) to train the model, the performance cannot be guaranteed, which is like the \textit{cold start} scenario. 
Moreover, we can find that the AUC values of \textit{NN} are very small while only using target labeled data, which may be due to the overfitting on a small number of samples.

\subsection{Analysis}
\color{black}{
\subsubsection{Ablation Study}
To analyse the separate contribution of \method, we compare \method with three variants of \method in this section, as listed below:
\begin{itemize}
    \item \textit{SSOT}: we remove Self-paced ensemble from \method.
    \item \textit{SPSSOT\_NC}: we do not consider intra-domain structure during transferring, i.e., $\beta = 0$ in Eq. \eqref{eq:loss_sum}.
    \item \textit{SPSSOT\_NG}: we delete the group entropic loss during training, i.e., $\lambda = 0$ in Eq.~\eqref{eq:loss_sum}.
\end{itemize}

The results are shown in Fig. \ref{fig:ablation}. 
As we can see, compared with the complete model, \textit{SSOT} is worse. This is because after removing the Self-paced ensemble, the datasets encounter a label imbalance that will result in the difficulty of modeling. 
\textit{SSPSOT\_NC} ignores the intra-domain structure with no consideration of the embedding distances in the hidden feature space; it is thus hard to find a good classification boundary.
What's more, \textit{SPSSOT\_NG} causes that the paired target unlabeled samples may come from different classes; this would lead to an ambiguous result.
In brief, the results indicate that each part of our model \method is necessary.

\subsubsection{Convergence}
To illustrate the convergence of \method, we evaluate the test AUCs of the transfer learning methods, \method and \textit{DeepJDOT}. 
The results are shown in Fig. \ref{fig:convergence}. 
It reveals that our model can achieve significantly better test AUCs only with a few iterations and keep relatively stable convergence performance.
On the task of Challenge $\to$  MIMIC, there are obvious changes in some iterations, like the 1000th and the 2000th iterations. 
That is because after every 1000 iterations, \method will resample from the majority data and continue training. 
With the increasing resampling times, the test performance will gradually become stable.

% feature visualization
\begin{figure*} 
    \centering
    \subfigure[Finetune (M$\to$C)]{
    \includegraphics[width=0.13\textwidth]{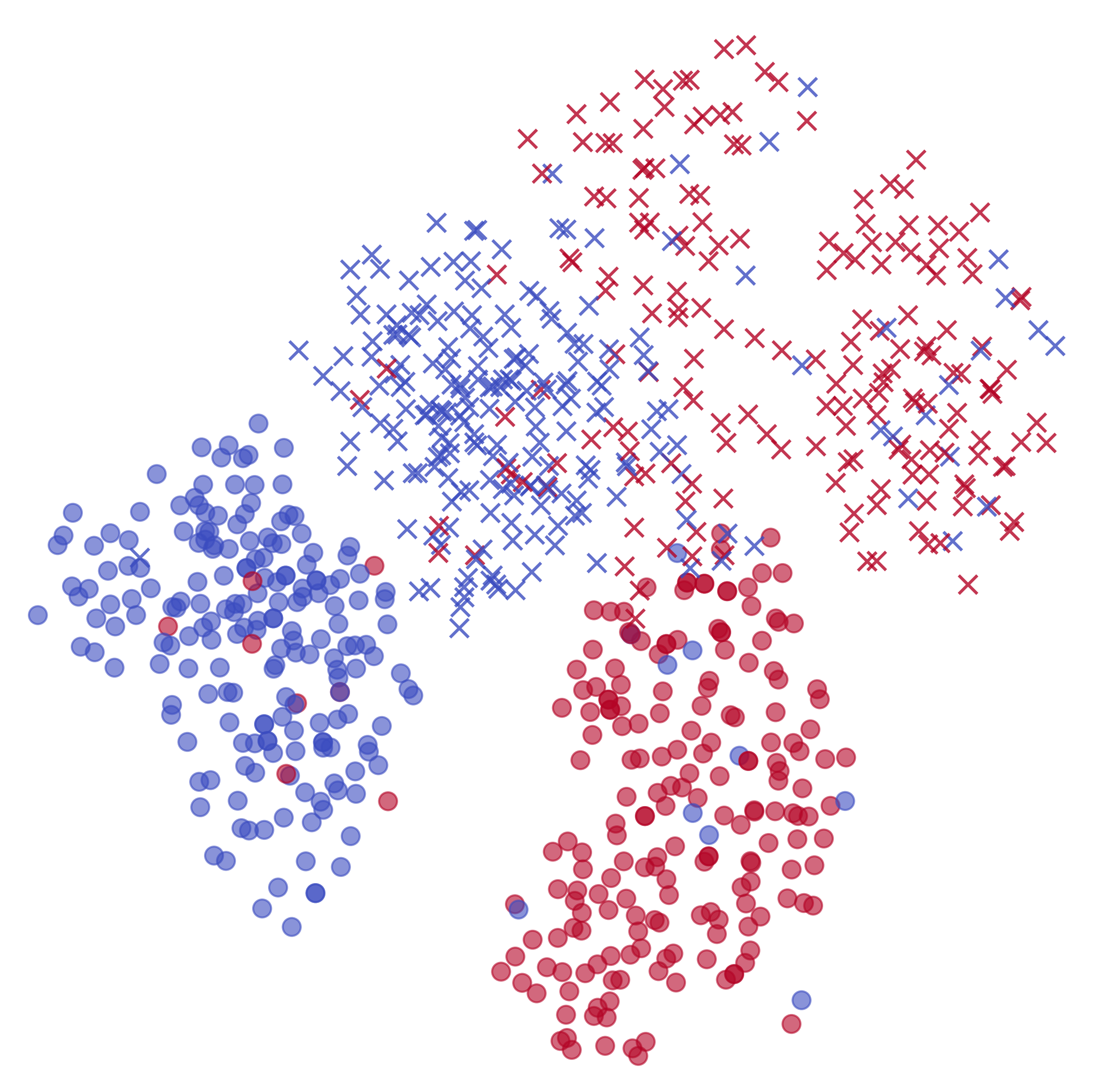}
    }
    \quad
    \subfigure[DeepJDOT(M$\to$C)]{
    \includegraphics[width=0.135\textwidth]{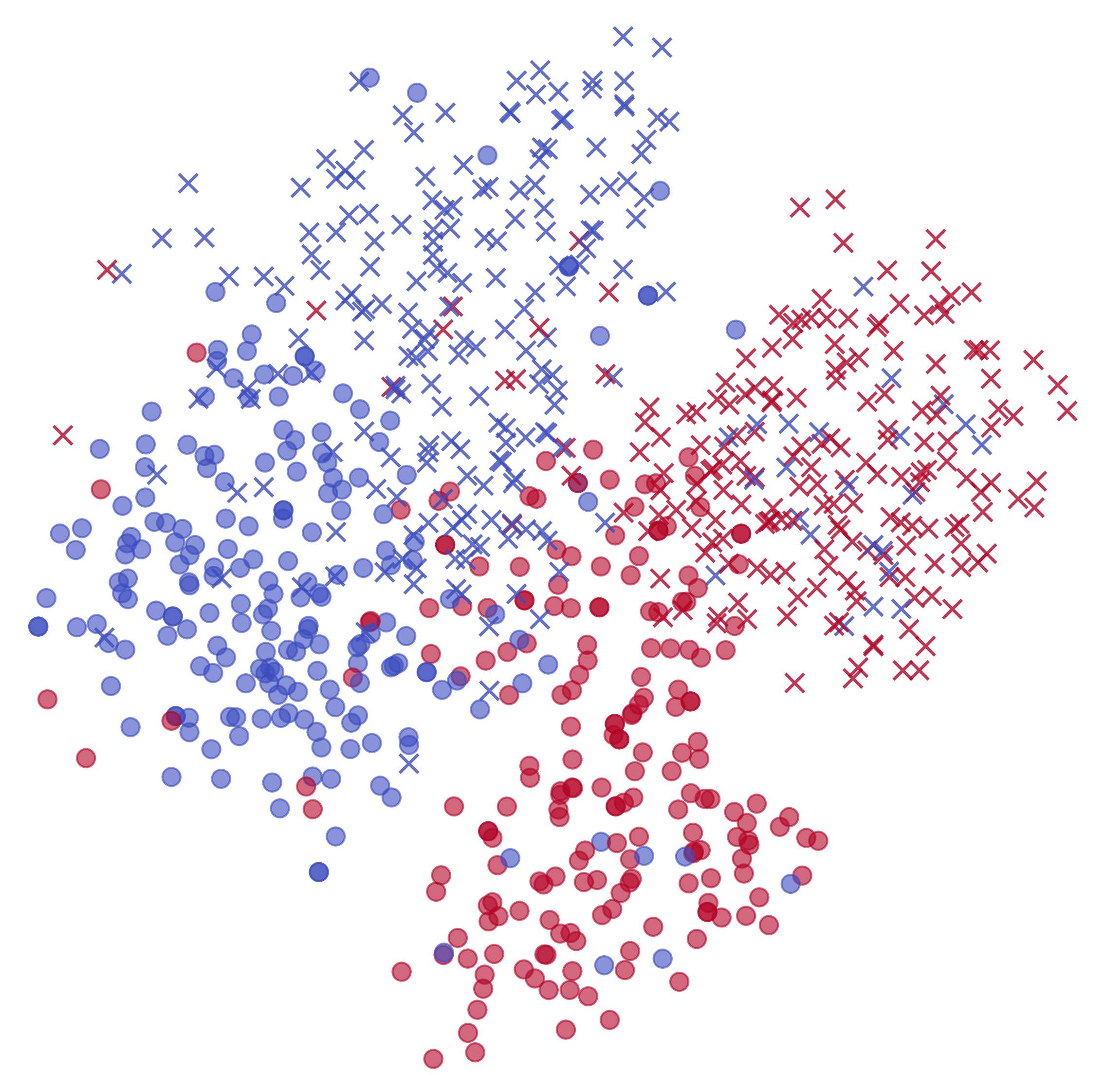}
    }
    \quad
    \subfigure[SPSSOT (M$\to$C)]{
    \includegraphics[width=0.13\textwidth]{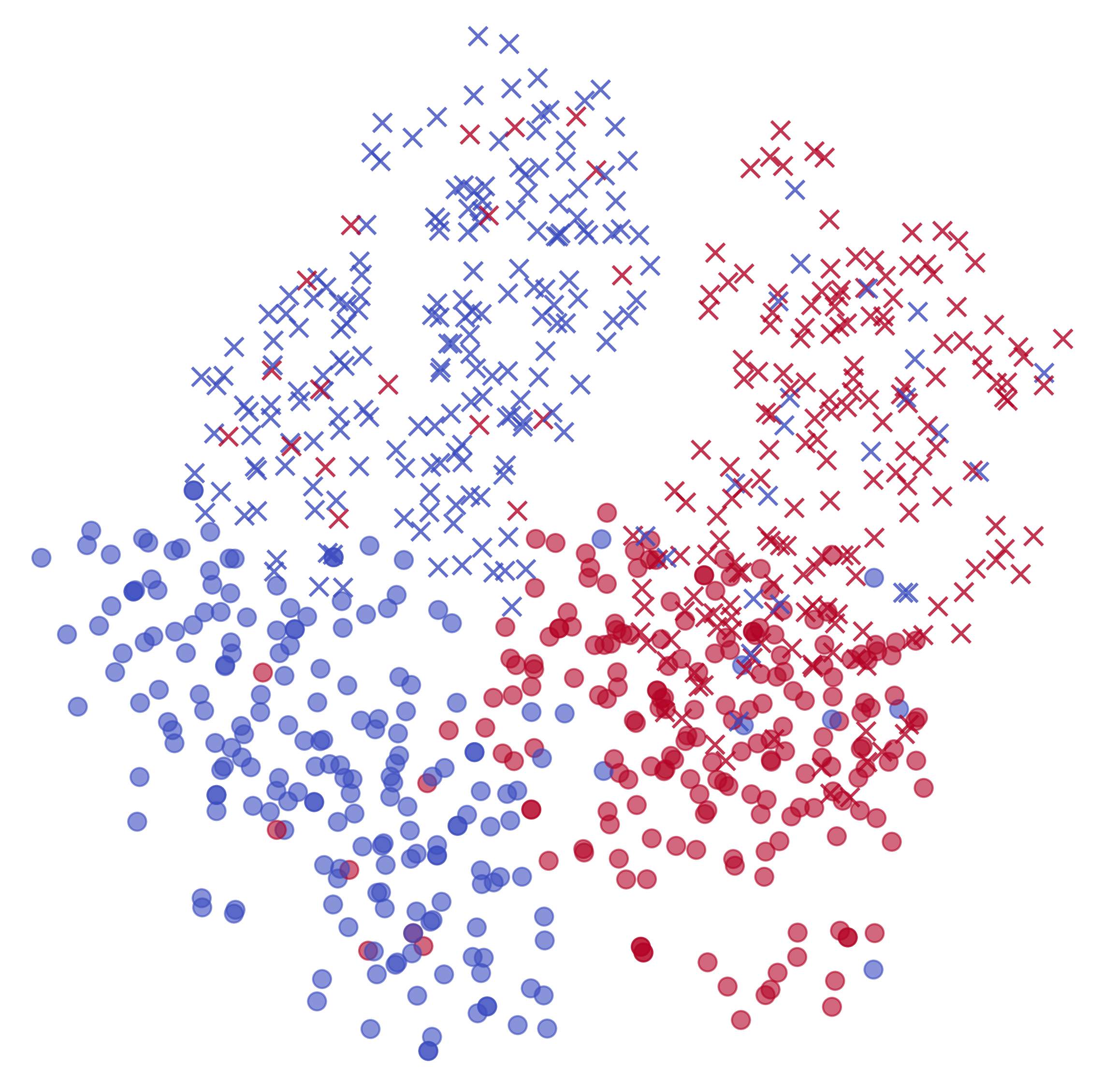}
    }
    \quad
    \subfigure[Finetune (C$\to$M)]{
    \includegraphics[width=0.13\textwidth]{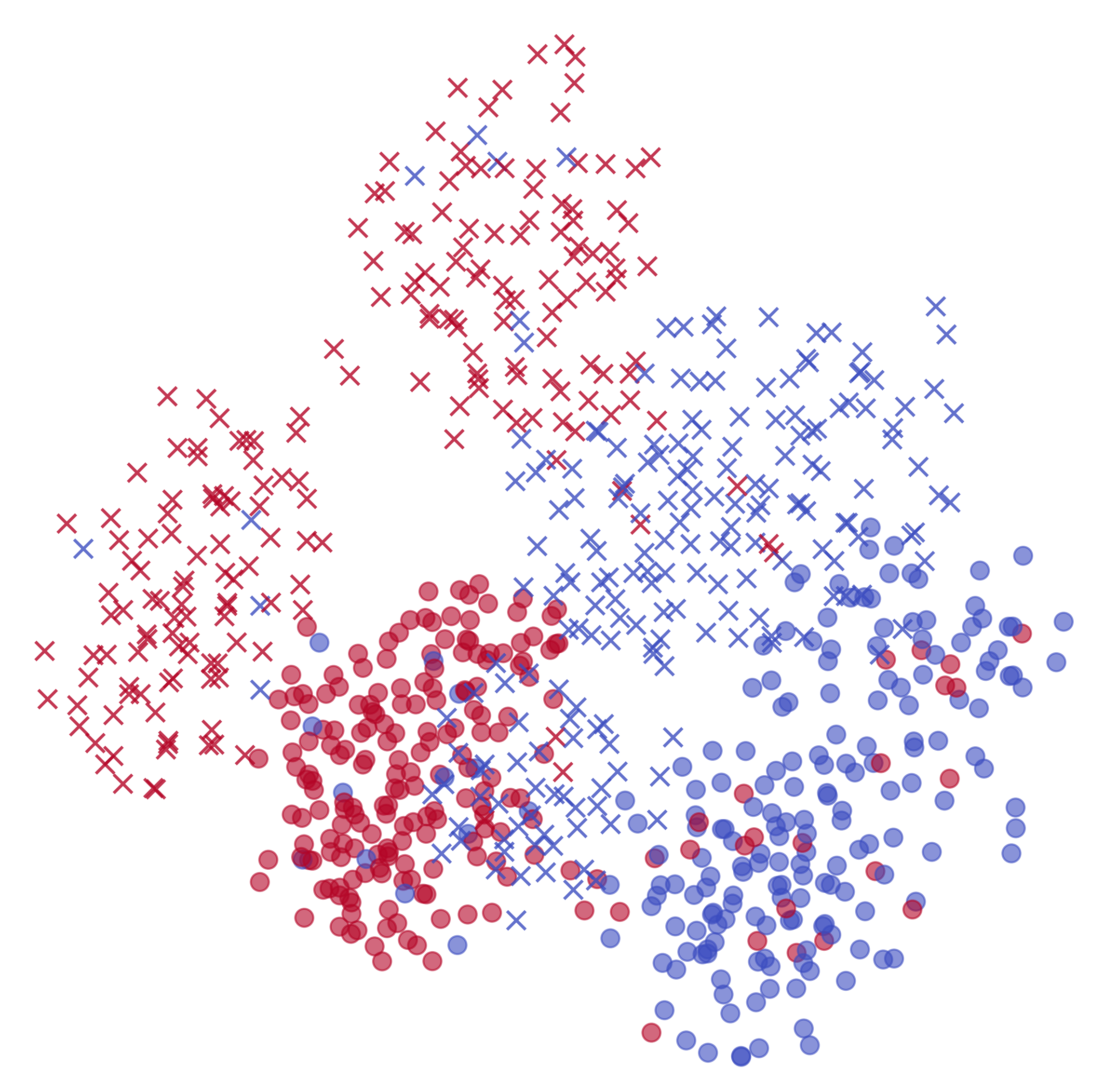}
    }
    \quad
    \subfigure[DeepJDOT(C$\to$M)]{
    \includegraphics[width=0.135\textwidth]{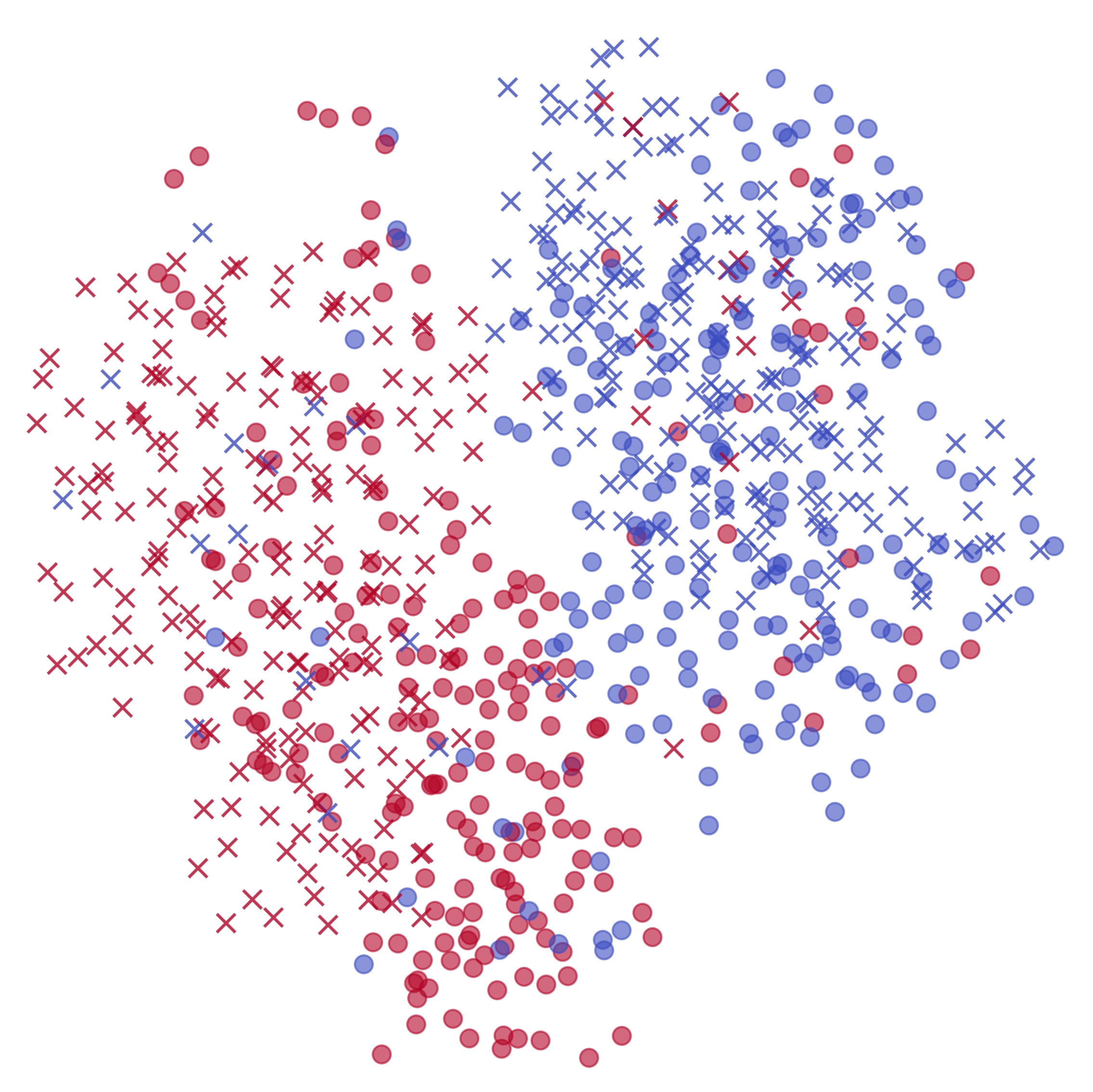}
    }
    \quad
    \subfigure[SPSSOT (C$\to$M)]{
    \includegraphics[width=0.13\textwidth]{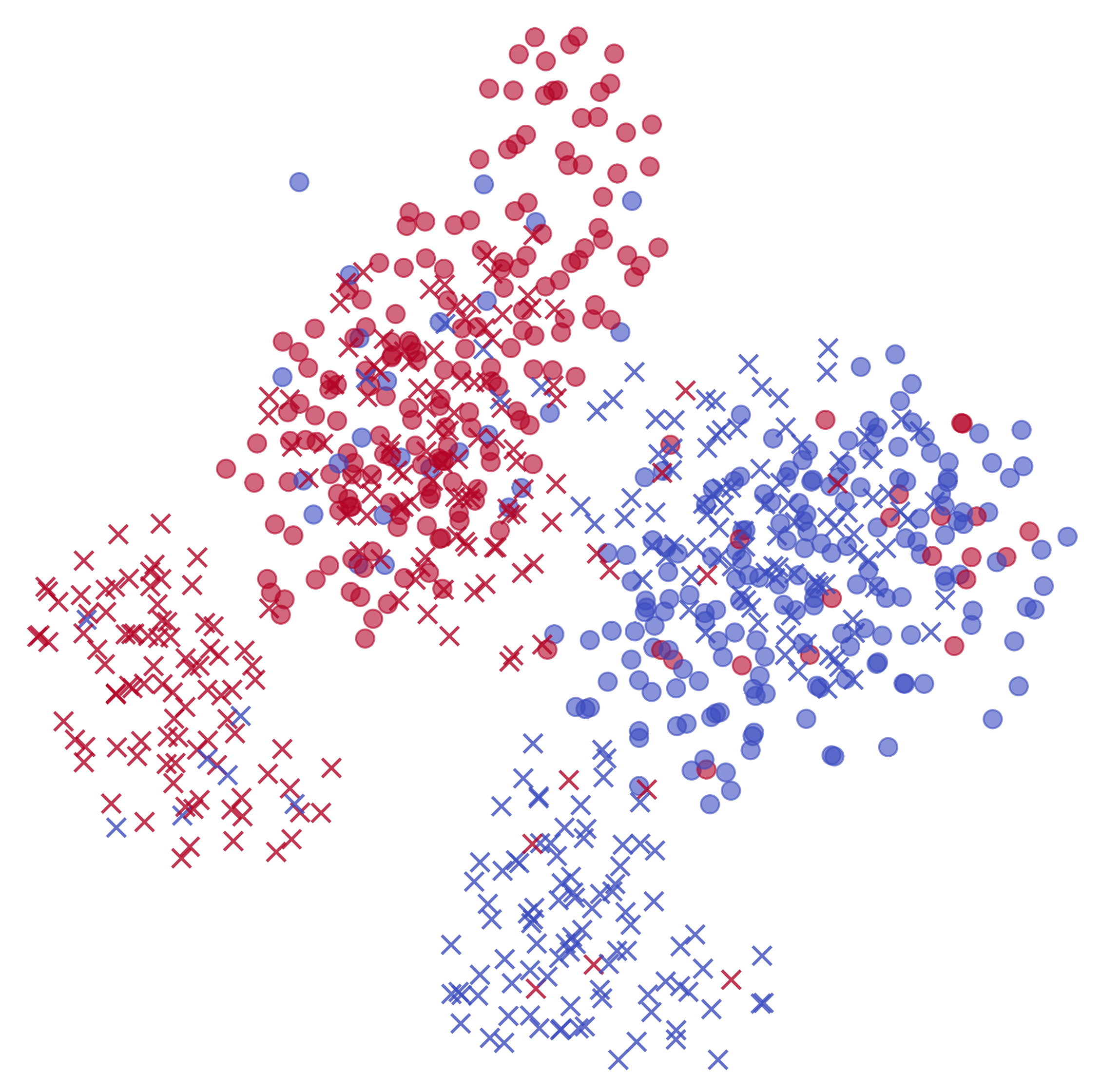}
    }
    \caption{Feature visualization. M $\to$ C means MIMIC $\to$ Challenge and C $\to$ M means Challenge $\to$ MIMIC. Different colors represent different classes (red: will have Sepsis, blue: will not have Sepsis), different shapes represent different domains (round: source domain, cross: target). \textit{Best viewed in color.}}
    \label{fig:feature_visualization}
\end{figure*}

\subsubsection{Sensitivity Analysis}
\label{sec:sensitivity}
\textbf{Labeled Percentage in Target Domain}. 
In experiments, we set 1\% of the target domain data to have labels by default. 
To further verify the stability of the method, we adjust the proportion of samples with known labels in the target domain to 0.5\%, 2\% and 4\%. 
In Fig. \ref{fig:label_percent}, we show the results of different target label percentages. 
It can be observed that  \method always performs the best in these percentages. 
It is reasonable that as the labeled percentage decreases, so does the models' performances. 
However, compared with the models that are trained with source data and target labeled data, \method keeps a relatively steady trend. 
This is an ideal situation for practical applications, which means that we can train a transfer learning model with acceptable performance by spending a small amount of cost to label a small amount of data.

\textbf{Sampling Percentage of Source Data}.
To validate the effect of different numbers of source domain samples, we conduct experiments on source data with different sample sizes.
Fig. \ref{fig:source_percentage} displays the results. 
We observe that the performance of all methods increases when using more source samples. 
At the same time, our \method approach consistently outperforms the other two baselines, which demonstrates the effectiveness of our proposed method for knowledge transfer.

\textbf{Hyper-parameter Sensitivity}. 
There exist four important hyper-parameters in the loss function of \method: the weight of optimal transport $\alpha$ in Eq. \eqref{eq:loss_lot}, the weight of the source data's classification loss $\theta_{s}$ in Eq. \eqref{eq:loss_lot},  the weight of discriminative centroid loss $\beta$, and the weight of group entropy $\lambda$ in Eq. \eqref{eq:loss_sum}. 
To test the stability of the performances of \method, we take a transfer scenerio, MIMIC $\to$ Challenge, as example to test different values of $\alpha$, $\theta_{s}$, $\beta$ and $\lambda$. 
The results are shown in Fig. \ref{fig:sensitivity}. 
Comparatively speaking, the model is not sensitive to all these parameters and the AUC just ranges from around 64 to 65.
According to the performance, we select the values of these parameters used in our experiments, i.e., $\alpha=0.05, \theta_{s} = 1, \beta = 0.15$ and $\lambda = 0.5$.
}

\subsubsection{Different Sepsis Early Detection Targets} 
Early Sepsis detection is potentially life-saving because doctors can treat earlier \cite{reyna2019early}. 
In the default setting, we predict Sepsis 6 hours before clinical diagnosis. 
Here, we add two early detection targets, 2 hours ahead and 4 hours ahead. The results are listed in Table \ref{tab:hours}. 
We can find that the performance is best when the advance time is 6 hours.
This is because when the advance time is short, the data imbalance will be exacerbated (i.e., the proportion of having Sepsis is decreased).

\begin{table}[t]
\centering
\caption{Results of Different Sepsis Early Detection Targets.}
\label{tab:hours}
\begin{tabular}{lcc}
\toprule
\textbf{Advance Time} & \textbf{MIMIC $\to$ Challenge} & \textbf{Challenge $\to$ MIMIC} \\ \hline            
\textit{2 hours}    & 63.53 $\pm$ 0.51   & 71.64 $\pm$ 0.71   \\
\textit{4 hours}    & 64.08 $\pm$ 0.35   & 72.30 $\pm$ 0.65    \\ 
\textit{6 hours(default)}   & 65.10 $\pm$ 0.24   & 76.05 $\pm$ 0.54  \\ 
\bottomrule
\end{tabular}
\end{table}

\subsubsection{Diverse Feature Generators} 
Considering the physiological indicators can be regarded as time-series data, we adapt the popular time series networks, LSTM \cite{hochreiter1997long} and GRU \cite{chung2014empirical}, as the feature generators of \method. 
As seen in Table \ref{table:rnn},  LSTM and GRU perform worse than NN (the default feature generator). 
Meanwhile, by fixing the feature generator (e.g., LSTM or GRU), \method consistently performs the best. This verifies the robustness of our method in applying different feature generators. 
We also concatenate the feature representations of NN and GRU (i.e., NN+GRU), but the results are still not as good as those of using only NN. 
These results inspire us that for exploiting time series models, perhaps more advanced feature engineering techniques are required.

\begin{table}[]
\centering
\caption{Results of Different Feature Generators in \method.}
\label{table:rnn}
\begin{tabular}{lccccc}
\toprule
 & \textbf{MIMIC $\to$ Challenge} & \textbf{Challenge $\to$ MIMIC} \\ \hline
\multicolumn{3}{l}{{\emph {\textbf{Source Only}}}}                       \\
\textit{NN}          & 59.24 $\pm$ 0.75  & 70.53 $\pm$ 1.34   \\
\textit{LSTM}        & 54.58 $\pm$ 0.50  & 68.26 $\pm$ 1.01 \\
\textit{GRU}         & 54.87 $\pm$ 0.97  & 69.72 $\pm$ 0.96  \\ 
\textit{NN+GRU}      & 57.37 $\pm$ 0.74  & 70.53 $\pm$ 0.93 \\ \hline
\multicolumn{3}{l}{{\emph {\textbf{Target Only}}}}                      \\
\textit{NN}          & 60.58 $\pm$ 0.14    & 61.92 $\pm$ 0.29 \\ 
\textit{LSTM}        & 54.74 $\pm$ 0.93   &  58.75 $\pm$ 1.29  \\   
\textit{GRU}         & 57.62 $\pm$ 0.59    & 57.29 $\pm$ 0.76 \\ 
\textit{NN+GRU}      & 59.35 $\pm$ 0.86    & 59.99 $\pm$ 0.54 \\ \hline
\multicolumn{3}{l}{{\emph {\textbf{Source \& Target Train Together}}}}    \\
\textit{NN}              & 60.81 $\pm$ 0.24 & 71.53 $\pm$ 0.09   \\
\textit{LSTM}            & 54.67 $\pm$ 1.05 & 68.83 $\pm$ 0.97   \\
\textit{GRU}             & 58.14 $\pm$ 1.10 & 70.56 $\pm$ 0.94   \\ 
\textit{NN+GRU}          & 59.35 $\pm$ 0.36 & 71.52 $\pm$ 0.48   \\ \hline
\multicolumn{3}{l}{{\emph {\textbf{Source \& Target Transfer}}}} \\     
\textit{Finetune} \textit{(NN)}    & 60.11 $\pm$ 0.73  & 71.62 $\pm$ 1.72  \\ 
\textit{Finetune} \textit{(LSTM)}  & 58.25 $\pm$ 0.77  & 70.03 $\pm$ 0.90   \\ 
\textit{Finetune} \textit{(GRU)}   & 59.60 $\pm$ 0.76  & 70.22 $\pm$ 1.00   \\ 
\textit{Finetune} \textit{(NN+GRU)}   & 60.05 $\pm$ 0.49  & 71.29 $\pm$ 1.26   \\
\method\textit{(NN)}             & \textbf{65.10 $\pm$ 0.24}  & \textbf{76.05 $\pm$ 0.54}   \\ 
\method\textit{(LSTM)}           & 60.45 $\pm$ 0.82  & 73.67 $\pm$ 0.62   \\ 
\method\textit{(GRU)}            & 62.09 $\pm$ 0.90  & 73.94 $\pm$ 0.50   \\ 
\method\textit{(NN+GRU)}         & 63.81 $\pm$ 0.51  & 75.18 $\pm$ 0.78   \\ 
\bottomrule
\end{tabular}
\end{table}

\color{black}{
\subsubsection{Feature visualization}
To show the feature transfer capability, we visualize the t-SNE embeddings \cite{van2008visualizing} of the hidden representation by \textit{Finetune}, \textit{DeepJDOT} and \method. 
Fig. \ref{fig:feature_visualization} (a) - \ref{fig:feature_visualization} (c) correspond to MIMIC $\to$ Challenge and Fig. \ref{fig:feature_visualization} (d) - \ref{fig:feature_visualization} (f) correspond to Challenge $\to$ MIMIC. 
In each sub-figure, different colors denote different categories (red: will have Sepsis, blue: will not have Sepsis), and different shapes denote different domains (round: source domain, cross: target). 
Fig. \ref{fig:feature_visualization}(a) and Fig. \ref{fig:feature_visualization}(d) display that the features learned by \textit{Finetune} for different domains are almost totally separated, i.e., points represented by different shapes in the same feature space are separated from each other.
% \update{And the classification boundaries of the two domains are completely different.}
Fig. \ref{fig:feature_visualization}(b) and Fig. \ref{fig:feature_visualization}(e) illustrate that though the domains can be aligned to a certain extend, the bad thing is some target samples are aligned to the source data with wrong classes, causing negative transfer. 
Note that, Fig. \ref{fig:feature_visualization}(c) and Fig. \ref{fig:feature_visualization}(f) show that the features generated by \method achieve better domain alignment with a clearer class boundary.
Specifically, when comparing Fig. \ref{fig:feature_visualization}(e) and Fig. \ref{fig:feature_visualization}(f), they both can blend the data from two domains well (i.e., the circled points and the crossed points are mixed). However, the classification boundary of \textit{DeepJDOT} is not as clear as that of \method. In particular, \textit{DeepJDOT}’s blue and red samples near the boundary are more interleaved than \method, which increases the likelihood that they will be incorrectly classified.
}
In a nutshell, the visualization results reveal that our proposal can match the complex structures of the source and target domains as well as maximize the margin between different classes. 

\section{Conclusion}
In this paper, we describe a new framework based on optimal transport and self-paced ensemble to solve the semi-supervised transfer learning problem for Sepsis early detection in the scenario that there is only little labeled data in the target domain (e.g. hospital). 
Empirical studies on real-world clinical datasets demonstrate the effectiveness of \method in aligning feature spaces and eliminating the influence of class imbalance. 
In fact, though \method is proposed for Sepsis early detection, it can be easily adapted for other transfer learning tasks.
The only requirement is choosing a suitable structure to extract deep features, e.g., CNNs for image identification \cite{xu2020reliable} and RNNs for time series prediction \cite{hochreiter1997long}. 

It is no doubt that there are still many problems to be solved. 
First, we only downsample from the labeled data to mitigate the effects of data imbalance. It is worthwhile to think about how to downsample from the target unlabeled data effectively to improve the accuracy of detection.
Moreover, we can further explore how to exploit time series models better.
After that, when there are private features in the source domain, it is hard to directly apply the optimal transport technique. Because feature similarity cannot be appropriately calculated between a source sample and a target sample. Therefore, it may require incorporating more transfer learning techniques, e.g., knowledge distillation \cite{ma2021distilling}.

\color{black}{Finally, privacy protection is an important issue that cannot be ignored when models are implemented in real-world applications. 
However, the constraints of privacy protection clauses often prevent data from being moved to the data center for unified storage and training. Federated learning \cite{yang2019federated} provides a new idea and still needs to be explored.
}

% \section*{Acknowledgment}
% This 

\bibliographystyle{IEEEtran}
\color{black}{\bibliography{reference.bib}}

% \newpage

\begin{appendix}
% \appendices
Because of the page limit, the full contents can be directly viewed at \url{https://ruiqingding.github.io/SPSSOT/},
including the following:
\begin{itemize}
    \item A. Training Time Consumption
    \item B. Synchronous Self-paced Downsampling
    \item C. Analysis of Outlier Disturbance
    \item D. Selection of $\rho$ in Label Adaptive Constraint
    \item E. Unmatched Features
\end{itemize}
\end{appendix}

\end{document}